\documentclass[nature,
,showpacs
,longbibliography
,superscriptaddress
]
{revtex4-2}

\usepackage{etoolbox}
\usepackage{graphicx,subfigure,bbm}

\usepackage{amsmath}
\usepackage{booktabs}  
\usepackage{placeins}
\usepackage{tabularx}
\usepackage{algorithm}
\usepackage{algpseudocode}
\usepackage{titletoc}
\usepackage{float}
\usepackage{xr-hyper}
\usepackage{nicefrac}        
\usepackage{amsthm}
\usepackage{amssymb}
\usepackage{array,bm}
\usepackage{graphicx}
\usepackage{fancyhdr}
\usepackage{color}
\usepackage{extarrows}
\usepackage{enumerate}
\usepackage{epstopdf}
\usepackage[colorlinks,
            linkcolor=black,
            citecolor=black,
            urlcolor=blue
            ]{hyperref}

\usepackage[utf8]{inputenc}
\usepackage{tikz}
\usetikzlibrary{shapes,arrows}
\usepackage{soul}
\newcommand{\fitfigure}[1]{%
    \includegraphics[width=\linewidth,height=0.82\textheight,keepaspectratio]{#1}%
}

\tikzstyle{decision} = [diamond, draw, fill=blue!20,
    text width=4.5em, text badly centered, node distance=3cm, inner sep=0pt]
\tikzstyle{block} = [rectangle, draw, fill=blue!20,
    text width=10em, text centered, rounded corners, minimum height=4em]
\tikzstyle{line} = [draw, -latex']
\tikzstyle{cloud} = [rectangle, draw,fill=red!20, node distance=7cm,
    minimum height=4em]

\newcommand{\be}{\begin{equation}}
\newcommand{\ee}{\end{equation}}
\newcommand{\bea}{\begin{eqnarray}}
\newcommand{\eea}{\end{eqnarray}}

\DeclareMathAlphabet{\varmathbb}{U}{bbold}{m}{n}

\usepackage{xcolor}

\usepackage{todonotes}

\usepackage{amsmath}
\usepackage{amsthm}
\usepackage{amssymb}
\usepackage{amsfonts}
\usepackage{array}
\usepackage{graphicx}
\usepackage{mathrsfs}
\usepackage{multirow}
\usepackage{booktabs}
\usepackage{siunitx}
\usepackage{float} 
\usepackage{subfigure} 
\newcommand{\beao}{\begin{eqnarray}}
\newcommand{\eeao}{\end{eqnarray}}

\begin{document}
\title{Kolmogorov-Arnold Reservoir Computing}

\author{Juntian Huang}
\affiliation{Institute of Fundamental and Frontier Sciences, University of Electronic Science and Technology of China, Chengdu 611731, China}

\author{J{\"u}rgen Kurths}
\affiliation{Potsdam Institute for Climate Impact Research, Potsdam 14412, Germany}
\affiliation{Department of Physics, Humboldt University Berlin, Berlin 12489, Germany}
\affiliation{\mbox{Research Institute of Intelligent Complex Systems, Fudan University, Shanghai 200433, China}}

\author{Ying Tang}
\email[Corresponding authors: ]{jamestang23@gmail.com}
\affiliation{Institute of Fundamental and Frontier Sciences, University of Electronic Science and Technology of China, Chengdu 611731, China}
\affiliation{School of Physics, University of Electronic Science and Technology of China, Chengdu 611731, China}
\affiliation{Key Laboratory of Quantum Physics and Photonic Quantum Information, Ministry of Education, University of Electronic Science and Technology of China, Chengdu 611731, China}
\affiliation{Non-classical Information Science Basic Discipline Research Center of Sichuan Province, University of Electronic Science and Technology of China, Chengdu 611731, China}

\begin{abstract}
Reservoir computing offers a lightweight framework for forecasting dynamical systems but may struggle to capture long-range dependencies due to limited representational capacity. Conventional reservoir computing recurrently uses fixed reservoirs with hyperparameter sensitivity, while the next generation reservoir computing removes recurrence at the cost of rapidly growing feature dimensions. Here, we develop Kolmogorov-Arnold Reservoir Computing (KARC), which replaces reservoirs with explicit basis-function expansions inspired by the Kolmogorov-Arnold representation theorem. We rigorously show that KARC is a lightweight design of Kolmogorov-Arnold networks (KANs), preserving the potential expressive capacity of KANs while admitting efficient closed-form training of reservoir computing. At comparable cost, KARC outperforms existing reservoir computing methods on challenging benchmarks including partial differential equations. It can also be integrated with generative diffusion models for facilitating  text-to-image generation. This work thus establishes a principled bridge between reservoir computing and KANs, yielding a unified framework for efficient dynamical forecasting and generative modeling.
\end{abstract}

\maketitle

\section{Introduction}
Data-driven forecasting of dynamical systems is a fundamental problem in computational science, with applications in climate modeling, fluid dynamics, finance and biological systems~\cite{TimeSeries,kovachki2023neural}. 
Deep learning models, including recurrent neural networks~\cite{RNN}, transformers~\cite{Attention} and neural operators~\cite{FNO,LNO}, have achieved strong performance in learning spatiotemporal dynamics and solution operators for partial differential equations (PDEs)~\cite{DeepONet,LNO,goswami2023physics,kovachki2024operator,liu2025architectures}. 
However, such models often require large training datasets, costly gradient-based optimization and substantial computational resources, limiting their applicability in data-scarce or resource-constrained scientific settings~\cite{TimeSeries,benidis2022deep}.
This motivates lightweight forecasting methods that retain predictive accuracy while reducing training cost and model complexity.

Reservoir computing (RC)~\cite{ESN,pathak2018model} offers such a lightweight paradigm for dynamical-system forecasting. 
In conventional RC, such as echo state networks~\cite{ESN} and liquid state machines~\cite{LSM,lin2025resistive}, a fixed recurrent reservoir maps inputs into a high-dimensional reservoir state space through nonlinear state updates, while only a linear readout is typically trained, usually by ridge regression.
This design achieves competitive accuracy at low training cost~\cite{zimmermann2018observing,xiong2019chaotic,fan2020long,PhysRevX.10.041037,lin2024learning,tan2024improving,li2024higher,han2026very,amann2026nonlinear}. 
Theoretical results further support this modeling paradigm, showing that RC-related architectures can approximate broad classes of fading-memory filters and input-output dynamical systems under suitable assumptions~\cite{grigoryeva2018universal,grigoryeva2018echo,gonon2019reservoir,hart2020embedding,gonon2021fading}. 
Nevertheless, RC remains sensitive to reservoir design and hyperparameters~\cite{RChyperparameter,ren2022global,yan2024emerging}, while its recurrent state updates introduce sequential dependencies that can hinder parallelization and scalability~\cite{lukosevicius2009reservoir,martin2017parallelizing}. 
The celebrated next generation reservoir computing (NG-RC)~\cite{NG-RC} removes explicit recurrence by reformulating RC as nonlinear vector autoregression with engineered polynomial features~\cite{NVAR,NG-RC}. 
This simplification reduces training complexity and facilitates parallel computation. 
However, the feature dimension in NG-RC grows rapidly with input size and expansion order,leading to a feature-dimensional bottleneck in high-dimensional systems~\cite{VolterraRC,cestnik2026next}.

To address these limitations, we develop Kolmogorov-Arnold reservoir computing (KARC), a framework inspired by the Kolmogorov-Arnold representation theorem~\cite{kolmogorov1957representations,KAN,KAN2}.
KARC represents the dynamical state through explicit univariate basis-function expansions, projecting time-delay coordinates onto basis functions such as Fourier, B-spline or Chebyshev functions. 
The resulting nonlinear feature representation is then combined with a linear readout, whose weights are trained efficiently by ridge regression. 
Similar closed-form or locally optimal regression strategies have recently been explored for efficient sequence modeling~\cite{von2025mesanet}.
In this way, KARC avoids the recurrent reservoir used in conventional RC while retaining the closed-form training advantage of RC. 
Moreover, by applying basis-function expansions to each delayed coordinate, KARC achieves linear feature scaling with the input dimension, thereby avoiding the rapid combinatorial growth of feature dimensionality encountered in NG-RC.

We further rigorously establish that KARC can be viewed as a lightweight realization of Kolmogorov-Arnold networks (KANs)~\cite{KAN,KAN2}. 
In KANs, multivariate mappings are represented through compositions of learnable univariate functions placed on network edges, and these edge functions are typically optimized by backpropagation. 
We show that, under specific functions, the KAN representation can be reduced to a fixed nonlinear feature map followed by a linear readout. This observation provides the basis for KARC: instead of learning all edge functions through iterative gradient-based optimization, KARC fixes the univariate basis functions and only trains the output weights in closed form. 
Therefore, KARC bridges KANs and RC by retaining a KAN-inspired representation while adopting the efficient readout-training paradigm of RC.

We evaluate the KARC framework through experiments on chaotic ordinary differential equations and high-dimensional PDE-governed systems.
Across the double-scroll system, the Kuramoto-Sivashinsky equation, and the shallow water equations, KARC achieves more accurate long-horizon predictions than existing reservoir-computing methods. Beyond conventional dynamical-system forecasting, KARC can serve as a feature-forecasting module for accelerating diffusion sampling, building on the recently proposed method Spectrum~\cite{Han2026Adaptive}. In addition to the Chebyshev basis functions used in Spectrum, KARC supports a general framework for feature forecasting with alternative basis, such as Fourier and B-spline functions, which perform comparably well. We further analyze the error bounds  associated with these basis functions. These results suggest that KARC offers an efficient and expressive framework for modeling complex dynamics, with potential applications in generative diffusion models.

\section{Results}

\begin{figure}[!htbp]
    \centering
    \fitfigure{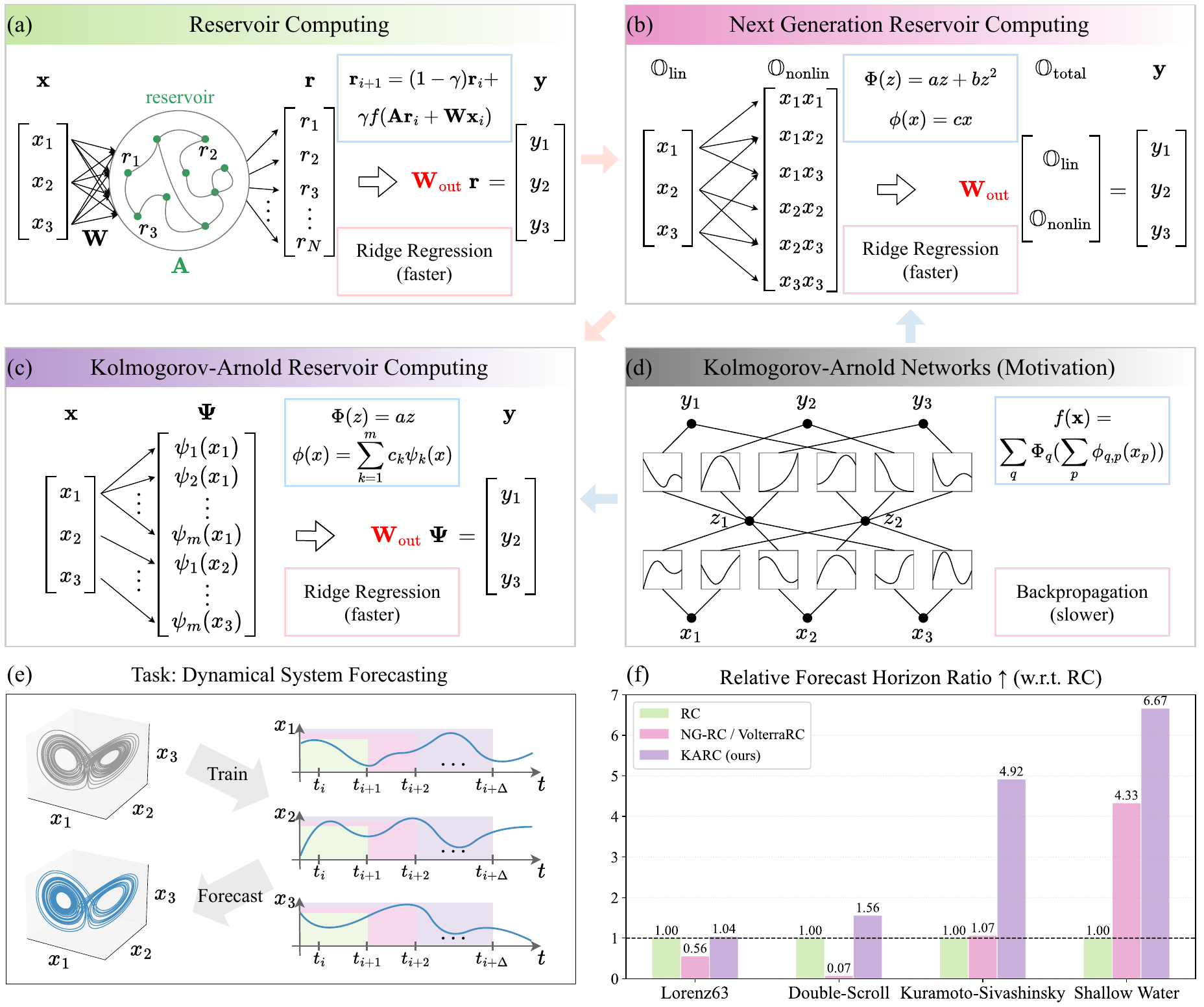}
    \caption{\textbf{Framework of Kolmogorov-Arnold Reservoir Computing.}
    (a) Conventional reservoir computing (RC) maps input states into a fixed recurrent reservoir and trains only a linear readout by ridge regression. 
    (b) Next generation reservoir computing (NG-RC) constructs polynomial features from the input state. 
    (c) Kolmogorov-Arnold reservoir computing (KARC) constructs structured nonlinear features from input coordinates through Kolmogorov-Arnold representation, as a lightweight form of Kolmogorov-Arnold networks (KANs) without backpropagation. Here, $\psi$ are prescribed basis functions such as Fourier, Chebyshev or B-spline functions. 
    (d) \(\phi\) and \(\Phi\) denote the inner and outer functions. The blue inset boxes show that NG-RC and KARC are KAN-related formulations under specific choices of \(\phi\) and \(\Phi\).
    (e) The dynamical-system forecasting task, with the coloured background indicating the forecast horizon of each reservoir-computing method.
    (f) Relative forecast horizon ratio with respect to RC on the Lorenz63 system, the double-scroll system, the Kuramoto-Sivashinsky equation, and the shallow water equations. NG-RC is used for the first two benchmarks, while VolterraRC~\cite{VolterraRC} with a similar form is used for the latter two to avoid the feature-dimensional explosion of NG-RC in high-dimensional systems.
}
    \label{fig:KARC_Overview}
\end{figure}

\subsection{Kolmogorov-Arnold Reservoir Computing}

To develop KARC, we first summarize the Kolmogorov-Arnold representation theorem, which states that any continuous multivariate function defined on a compact domain can be represented by a finite composition of univariate functions and addition. 
Specifically, for any continuous function \(f:[0,1]^n\to\mathbb{R}\), there exist continuous
univariate functions \(\phi_{q,p}:[0,1]\to\mathbb{R}\) and \(\Phi_q:\mathbb{R}\to\mathbb{R}\), with \(q=1,\ldots,2n+1\) and
\(p=1,\ldots,n\), such that
\begin{equation}
    f(x_1,\ldots,x_n)
    =
    \sum_{q=1}^{2n+1}
    \Phi_q
    \left(
    \sum_{p=1}^{n}
    \phi_{q,p}(x_p)
    \right).
\end{equation}
This theorem provides a theoretical foundation for representing continuous multivariate functions through compositions of one-dimensional functions.

KANs~\cite{KAN,KAN2} build on the Kolmogorov-Arnold representation theorem by assigning learnable univariate functions to network edges, which are typically trained by gradient-based optimization. 
Here, we observe that if these univariate functions are expanded over a fixed function dictionary rather than optimized as learnable edge functions, the resulting representation can be written as a fixed nonlinear feature map followed by a linear readout.
This observation naturally connects the Kolmogorov-Arnold representation with the training paradigm of RC, where nonlinear features are constructed first and only the output layer is trained.
Based on this idea, we propose KARC, which combines a univariate basis-function representation with the closed-form readout training of reservoir computing.

To obtain a computationally efficient model that admits closed-form ridge-regression training, we consider a linearized form of the Kolmogorov-Arnold representation by assuming that the outer functions are linear:
\begin{equation}
    \Phi_q(z) = a_q z .
\end{equation}
Each unknown univariate inner function \(\phi_{q,p}\) is then approximated by a finite expansion over a prescribed set of basis functions \(\{\psi_j\}_{j=1}^{m}\):
\begin{align}
    \phi_{q,p}(x_p)
    \approx
    \sum_{j=1}^{m}
    c_{q,p,j}\psi_j(x_p).
\end{align}
Substituting this expansion into the linearized representation gives
\begin{align}
    f(x_1,\ldots,x_n)
    &\approx
    \sum_{q=1}^{2n+1}
    a_q
    \left(
        \sum_{p=1}^{n}
        \sum_{j=1}^{m}
        c_{q,p,j}\psi_j(x_p)
    \right) \\
    &=
    \sum_{p=1}^{n}
    \sum_{j=1}^{m}
    \left(
        \sum_{q=1}^{2n+1}
        a_q c_{q,p,j}
    \right)
    \psi_j(x_p).
\end{align}
By absorbing the coefficients associated with the outer and inner functions into
a single readout coefficient,
\begin{equation}
    w_{p,j}
    =
    \sum_{q=1}^{2n+1}
    a_q c_{q,p,j},
\end{equation}
we obtain the KARC approximation:
\begin{equation}
    f(x_1,\ldots,x_n)
    \approx
    \sum_{p=1}^{n}
    \sum_{j=1}^{m}
    w_{p,j}\psi_j(x_p).
\end{equation}
Equivalently, by introducing the feature vector
\begin{equation}
    \mathbf{\Psi}(\mathbf{x})
    =
    \big[
    \psi_1(x_1),\ldots,\psi_m(x_1),
    \ldots,
    \psi_1(x_n),\ldots,\psi_m(x_n)
    \big]^\top ,
\end{equation}
the model can be written compactly as
\begin{equation}
    \widehat{f}(\mathbf{x})
    =
    \mathbf{W}_{\mathrm{out}} \mathbf{\Psi}(\mathbf{x}).
\end{equation}

To incorporate temporal information, we construct a time-delay embedding from the observed trajectory \(\{\mathbf{u}_i\}_{i=1}^{T}\), where the delay-embedded state is defined as 
\begin{equation} 
\label{equ:delayed_embedding}
    \mathbf{x}_i = \mathbf{u}_i \oplus \mathbf{u}_{i-1} \oplus \cdots \oplus \mathbf{u}_{i-k+1},
\end{equation}
with \(k\) denoting the delay length and \(\oplus\) denoting vector concatenation.
The basis-function feature map is then constructed from \(\mathbf{x}_i\), enabling the model to use both the current observation and its historical states for forecasting. 
The resulting feature vector is used for one-step prediction through a linear readout,
\begin{equation} 
    \widehat{\mathbf{u}}_{i+1} = \mathbf{W}_{\mathrm{out}} \mathbf{\Psi}(\mathbf{x}_i). 
\end{equation} 
Given the feature matrix 
\begin{equation} 
    \mathbf{H} = \left[ \mathbf{\Psi}(\mathbf{x}_k), \mathbf{\Psi}(\mathbf{x}_{k+1}), \ldots, \mathbf{\Psi}(\mathbf{x}_T) \right], 
\end{equation} 
and the corresponding target matrix 
\begin{equation} 
    \mathbf{Y} = \left[ \mathbf{u}_{k+1}, \mathbf{u}_{k+2}, \ldots, \mathbf{u}_{T+1} \right], 
\end{equation} 
the output weights are obtained by ridge regression:
\begin{equation}
    \mathbf{W}_{\mathrm{out}} = \mathbf{Y}\mathbf{H}^{\top} \left( \mathbf{H}\mathbf{H}^{\top} + \lambda \mathbf{I} \right)^{-1}, 
\end{equation} 
where \(\lambda\) is the ridge regularization coefficient. Therefore, KARC can be viewed as a reservoir-computing realization of the Kolmogorov-Arnold representation: it preserves the efficient closed-form readout training of RC while retaining the univariate-function representation principle of KANs.

Higher-order extensions of KARC are described in detail in the Methods section. 
These extensions are introduced to further improve the nonlinear representational capacity of the model by incorporating interactions among univariate basis responses. 
However, this increased expressiveness comes with a larger feature dimension and higher computational cost. 
Therefore, in practice, the maximum order of KARC should be selected to balance nonlinear representational capacity and training efficiency.

\subsection{Applications}

We evaluate the KARC framework on a range of dynamical systems, including both chaotic systems and high-dimensional PDE-governed systems. 
Unless otherwise stated, KARC is implemented with Fourier basis functions in the following experiments, while results with other basis functions are reported in the Supplementary Information~V.
Since KARC, RC, and NG-RC exhibit comparable forecasting performance on the Lorenz63 system, we report the corresponding results in the Supplementary Information~VII.A and focus the main text on more challenging benchmarks. 
For high-dimensional PDE-governed systems, we replace NG-RC with VolterraRC~\cite{VolterraRC,VolterraSeries}, since NG-RC suffers from feature-dimensional explosion in these settings.
Unless otherwise stated, RC, NG-RC, and KARC are evaluated on an NVIDIA H100 GPU server, whereas VolterraRC is evaluated on an AMD Ryzen Threadripper PRO 7975WX 32-Core CPU, as the original implementation of VolterraRC is CPU-based without GPU-compatible version available.

\subsubsection{Double-Scroll System}

\begin{figure}[!htbp]
    \centering
    \fitfigure{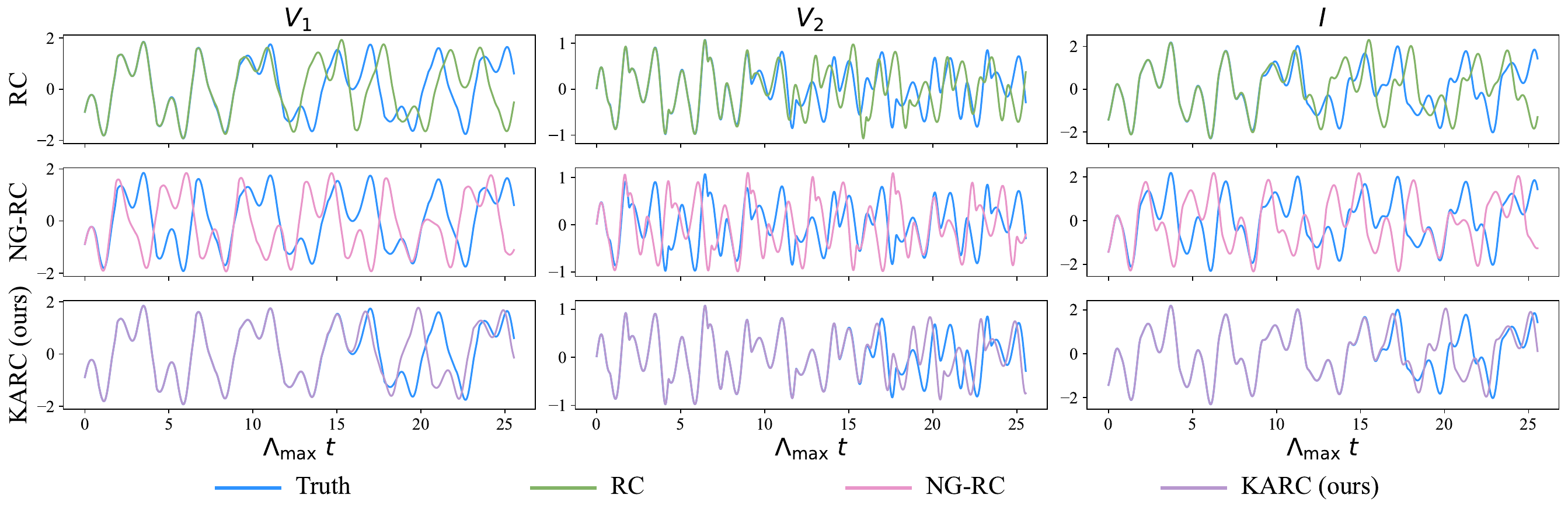}
    \caption{\textbf{Forecasting performance of RC, NG-RC, and KARC on the double-scroll system.} Rows correspond to different models, and columns correspond to the three state variables of the double-scroll system. $\Lambda_{\max}$ denotes the largest Lyapunov exponent, and one unit on the horizontal axis represents one Lyapunov time.}
    \label{fig:DoubleScroll_Trajectory}
\end{figure}

\begin{table}[!htbp]
    \centering
    \setlength{\tabcolsep}{10pt}
    \renewcommand{\arraystretch}{1.25}
    \begin{tabular}{lcccc}
        \toprule
        Model 
        & Dimension
        & Total Time (s) \(\downarrow\)
        & NRMSE$@$1LT \(\downarrow\)  
        & VPT ($\epsilon=0.1$) [LT] \(\uparrow\)\\
        \midrule
        RC & $3000$ & $0.357$ & $2.133 \times 10^{-2}$ & $10.720 $ \\
        NG-RC & $63$ & $0.653$ & $1.587 \times 10^{-1}$ & $0.800$ \\
        KARC (ours) & $1891$ & $0.120$ & $5.293 \times 10^{-4}$ & $16.736$ \\
        \bottomrule
    \end{tabular}
    \caption{\textbf{Quantitative comparison of forecasting performance on the double-scroll system.}
    Dimension denotes the feature dimension of each model, namely the dimension of the reservoir state 、、、\(\mathbf{r}\) in RC, the total feature vector \(\mathbb{O}_{\mathrm{total}}\) in NG-RC and the KARC feature vector \(\boldsymbol{\Psi}\).
    The total time includes both readout training and autonomous rolling prediction.
    We define the cumulative normalized prediction error as
    $e(t)=\sqrt{\mathrm{mean}[(\widehat{\mathbf{Y}}_{1:t}-\mathbf{Y}_{1:t})^2]/\mathrm{var}(\mathbf{Y}_{1:t})}$.
    NRMSE$@$1LT is defined as $e(T_{\mathrm{L}})$, where $T_{\mathrm{L}}$ denotes the number of time steps within one Lyapunov time.
    The valid prediction time (VPT) is defined as
    $\mathrm{VPT}=\inf\{t>0:e(t)>\epsilon\}$, with $\epsilon=0.1$.}
    \label{tab:doublescroll_comparison}
\end{table}

We first evaluate KARC on the double-scroll system, a canonical nonlinear electronic circuit with chaotic dynamics. 
Its dimensionless governing equations are
\begin{align}
\dot{V}_1 &= V_1/R_1 - \Delta V/R_2 - 2 I_r\,\sinh(\beta \Delta V), \\
\dot{V}_2 &= \Delta V/R_2 + 2 I_r\,\sinh(\beta \Delta V) - I, \\
\dot{I} &= V_2 - R_4 I,
\end{align}
in the chaotic regime, where $\Delta V = V_1 - V_2$. 
We set $R_1=1.2$, $R_2=3.44$, $R_4=0.193$, $\beta=11.6$, and $I_r=2.25\times10^{-5}$, yielding a Lyapunov time of approximately $7.81$. 
The training trajectory contains $4{,}000$ data points sampled at four observations per unit time.

All models are trained for one-step prediction and evaluated by autonomous rollout: RC uses a reservoir dimension of \(3000\), NG-RC adopts a third-order polynomial expansion and KARC uses a second-order Fourier basis-function expansion.
In Fig.~\ref{fig:DoubleScroll_Trajectory}, RC follows the overall trajectory for approximately \(9\) Lyapunov times before visible phase and amplitude errors emerge, whereas NG-RC rapidly departs from the reference trajectory. 
In contrast, KARC remains close to the reference dynamics for a substantially longer horizon, remaining accurate for approximately \(15\) Lyapunov times and better preserving the oscillatory structure of the double-scroll system. 
This qualitative behavior is consistent with the quantitative results in Table~\ref{tab:doublescroll_comparison}, where KARC achieves the lowest NRMSE over the first Lyapunov time, reducing the error by more than one order of magnitude relative to both RC and NG-RC.

Although KARC has a larger feature dimension than NG-RC in this experiment, it only requires second-order feature construction, whereas NG-RC uses a third-order nonlinear expansion. 
As a result, KARC is much faster than NG-RC during the feature-construction stage and achieves the shortest overall total time among the compared methods. 
Moreover, because the double-scroll system is only three-dimensional, the second-order KARC expansion does not introduce prohibitive dimensional growth. 
These results suggest that, for low-dimensional chaotic systems, KARC provides a richer basis-function representation while keeping the computational cost acceptable.

\subsubsection{Kuramoto-Sivashinsky Equation}

\begin{figure}[!htbp]
    \centering
    \fitfigure{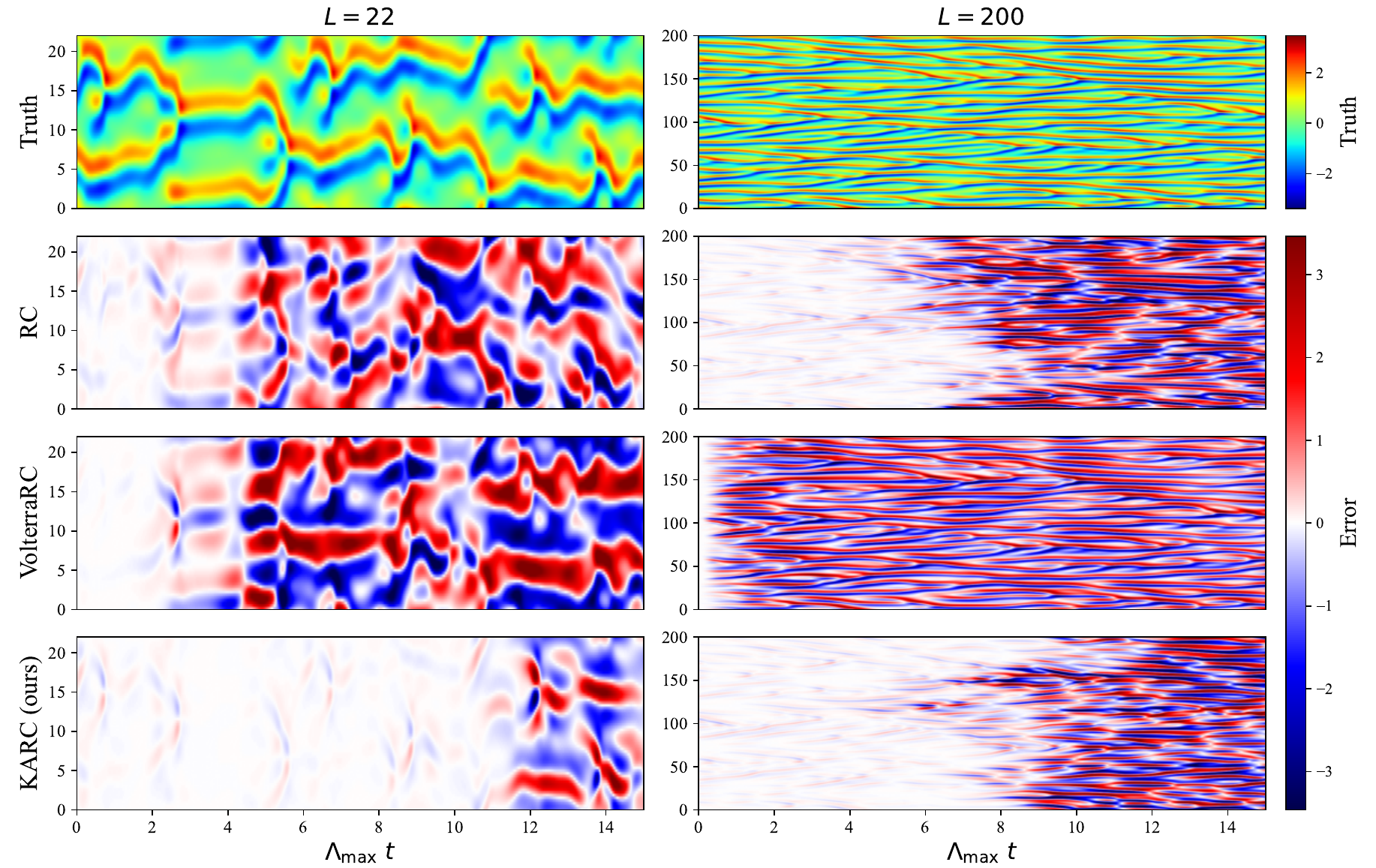}
    \caption{\textbf{Spatiotemporal forecasting performance on the Kuramoto-Sivashinsky equation.}
    The left (right) column corresponds to the \(L=22\) (\(L=200\)) setting, where the spatial domain is discretized into \(64\) (\(512\)) grid points.
    In the right column, the RC baseline is evaluated using a parallelized prediction scheme~\cite{pathak2018model} with 64 local reservoirs, and RC, KARC are evaluated on CPU due to GPU memory constraints. 
    In both settings, the first row shows the ground-truth solution, and the second to fourth rows show the relative errors of RC, VolterraRC, and KARC, respectively. 
    Here, \(\Lambda_{\max}\) denotes the largest Lyapunov exponent, and one unit on the horizontal axis corresponds to one Lyapunov time.}
    \label{fig:KS_Comparison_Figure}
\end{figure}

We next consider the Kuramoto-Sivashinsky (KS) equation, a standard benchmark for spatiotemporal chaos: 
\begin{equation} 
u_t + u u_x + u_{xx} + u_{xxxx} = 0, 
\end{equation} 
where \(u(x,t)\) is a scalar field defined on \(x\in[0,L]\) with periodic boundary conditions. 
The coupling between nonlinear advection, long-wavelength instability, and high-order dissipation generates irregular multiscale dynamics, making long-horizon forecasting substantially more challenging than the low-dimensional chaotic systems considered above. 
In this work, we consider two domain sizes, \(L=22\) and \(L=200\), representing a moderately chaotic regime and a large-scale spatiotemporal chaotic regime, respectively.
For the \(L=22\) and \(L=200\) settings, the training datasets consist of \(40{,}000\) and \(80{,}000\) data points, respectively, both sampled at a rate of four observations per unit time.

\begin{table}[!htbp]
    \centering
    \setlength{\tabcolsep}{10pt}
    \renewcommand{\arraystretch}{1.25}
    \begin{tabular}{lcccc}
        \toprule
        Model & Dimension & Total Time (s) \(\downarrow\) & NRMSE$@$1LT \(\downarrow\)& VPT ($\epsilon=0.1$) [LT] \(\uparrow\)\\
        \midrule
        RC & $3000$ & $2.82$ & $5.581 \times 10^{-2}$ & $2.413$ \\
        VolterraRC & $40000$ & $978.11$ & $1.589 \times 10^{-2}$ & $2.575$ \\ 
        KARC (ours) & $12801$ & $0.79$ & $6.301 \times 10^{-4}$ & $11.875$ \\
        \bottomrule
    \end{tabular}
    \caption{\textbf{Comparison of forecasting performance on the Kuramoto-Sivashinsky equation with \(L=22\).} The evaluation metrics are defined in the same way as in the double-scroll experiment. In addition, the total time of VolterraRC is substantially higher than that of the other models because VolterraRC is implemented on CPU, whereas the other models are run on GPU.}
    \label{tab:ks_comparison}
\end{table}

Fig.~\ref{fig:KS_Comparison_Figure} compares the forecasting performance of the considered models on the KS equation with \(L=22\) and \(L=200\). 
For the \(L=22\) setting, RC and VolterraRC capture the dominant spatiotemporal structures for approximately 5 Lyapunov times, after which their predictions gradually deviate from the ground truth.
In contrast, KARC maintains accurate spatiotemporal evolution for approximately 11 Lyapunov times, indicating a substantially longer forecasting horizon in this regime.
Table~\ref{tab:ks_comparison} further confirms that, although KARC uses a larger feature dimension than RC, it achieves lower total computational time and substantially higher forecasting accuracy.
Specifically, KARC reduces the NRMSE by nearly two orders of magnitude compared with RC and extends the VPT from about \(2.4\) to \(11.9\) Lyapunov times.

In the larger-domain setting \(L=200\), the RC baseline is implemented in a parallelized form using 64 local reservoirs, following the large-scale KS forecasting strategy in Ref.~\cite{pathak2018model}.
With this parallelized prediction scheme, RC maintains accurate forecasts for approximately \(7\) Lyapunov times, whereas VolterraRC exhibits large errors from the early stage of the autonomous rollout.
KARC maintains a low prediction error over the first \(9\) Lyapunov times, achieving a comparable or longer forecasting horizon than parallelized RC. 
This result is notable because KARC uses a single global feature representation, without explicitly partitioning the spatial domain or training multiple local reservoirs.

\subsubsection{Shallow Water Equations}

\begin{figure}[!htbp]
    \centering
    \fitfigure{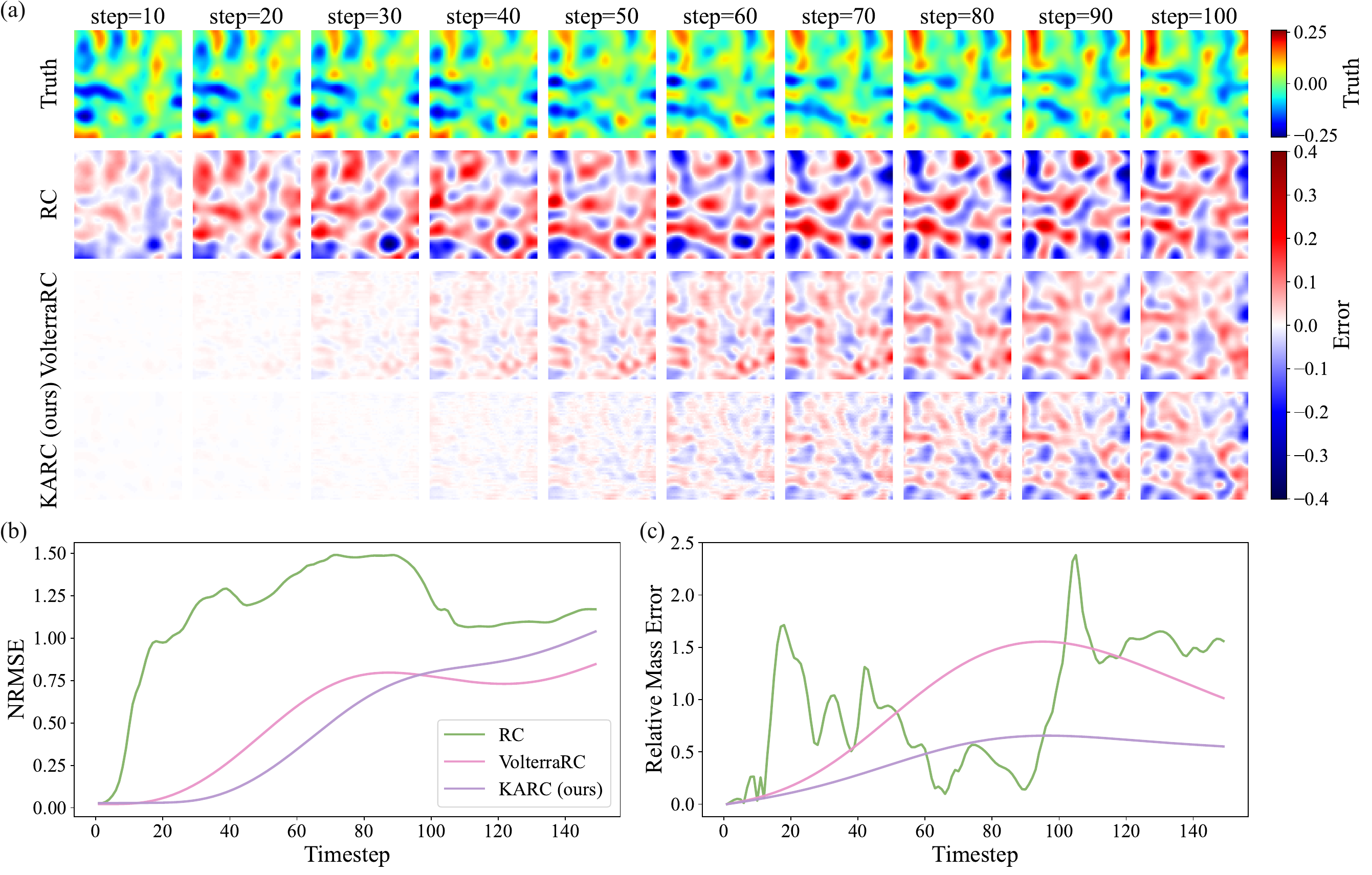}
    \caption{\textbf{Forecasting performance on the two-dimensional shallow water equations.} 
    (a) Reference solution (top row) and corresponding forecasts produced by RC, VolterraRC, and KARC at selected forecast steps from the same initial condition. 
    (b) Normalized root-mean-square error (NRMSE) as a function of forecast time step.
    (c) Relative mass error, defined as the deviation of the predicted total mass from its initial value, over the same rollout horizon.}
    \label{fig:SWE_Figure}
\end{figure}

We next consider the two-dimensional shallow water equations (SWE)~\cite{Vallis2017Atmospheric}, a standard model of geophysical fluid dynamics that describes the evolution of a thin fluid layer under gravity and rotation. 
In this work, we focus on a forced-dissipative rotating shallow-water regime, where the dynamics are driven by external wind stress and mass forcing, while Coriolis effects, gravity waves, and linear damping jointly shape the spatiotemporal evolution:
\begin{align}
    &\frac{\partial u}{\partial t} - f v = -g \frac{\partial \eta}{\partial x} + \frac{\tau_x}{\rho_0 H} - \kappa u, \\
    &\frac{\partial v}{\partial t} + f u = -g \frac{\partial \eta}{\partial y} + \frac{\tau_y}{\rho_0 H} - \kappa v, \\
    &\frac{\partial \eta}{\partial t} + \frac{\partial[(\eta+H)u]}{\partial x} + \frac{\partial[(\eta+H)v]}{\partial y} = \sigma - w ,
\end{align}
where $u$ and $v$ denote the horizontal velocity components and $\eta$ is the free-surface displacement. 
The parameters $f$, $g$, $\rho_0$, $H$, $\tau_x$, $\tau_y$, and $\kappa$ denote the Coriolis parameter, gravitational acceleration, reference water density, mean fluid depth, wind-stress forcing in the $x$ and $y$ directions, and the linear damping coefficient, respectively. 
The terms $\sigma$ and $w$ denote source and sink contributions in the mass-conservation equation. 

\begin{table}[!htbp]
    \centering
    \setlength{\tabcolsep}{10pt}
    \renewcommand{\arraystretch}{1.25}
    \begin{tabular}{lcccc}
        \toprule
        Model & Dimension & Total Time (s) \(\downarrow\) & NRMSE$@$10steps \(\downarrow\) &  VPT ($\epsilon=0.1$) [step] \(\uparrow\) \\
        \midrule
        RC & $5000$ & $0.83$ & $2.223 \times 10^{-1}$ & $6$ \\
        VolterraRC & $3000$ & $36.83$ & $2.335\times 10^{-2}$ & $26$ \\ 
        KARC (ours) & $98305$ & $0.47$ & $2.917 \times 10^{-2}$ & $40$ \\
        \bottomrule
    \end{tabular}
    \caption{\textbf{Quantitative comparison of forecasting performance on the shallow water equations.} 
    Model dimension and total time are reported using the same definitions as in the preceding experiments. 
    NRMSE is computed over the first ten forecasting steps. 
    The VPT denotes the first forecasting step at which the prediction error exceeds $\epsilon=0.1$; therefore, its unit is the discrete time step rather than the Lyapunov time.}
    \label{tab:SWE_comparison}
\end{table}

To represent large-scale ocean dynamics, we use a square computational domain with characteristic length on the order of $10^6$ m. 
Specifically, the spatial domain is defined as $L_x=L_y=10^6$ and discretized on a $64\times64$ grid. We set $H=100.0$, $g=9.81$, and $\rho_0=1024.0$. 
The Coriolis parameter is modeled as $f=f_0+\beta y$, with $f_0=1\times10^{-4}$ and $\beta=2\times10^{-11}$. 
The wind-stress forcing terms are set to zero, that is, \(\tau_x=\tau_y=0\).
Linear friction is included with a constant coefficient \(\kappa = 1/(5\times24\times3600)\), corresponding to a damping timescale of five days.
Source and sink terms are not included, that is, $\sigma=w=0$.
The training trajectory contains $3000$ time steps. 
The time step \(dt\) is chosen according to the Courant-Friedrichs-Lewy condition:
\begin{align}
    d x = \frac{L_x}{N_x-1}, \qquad
    d y = \frac{L_y}{N_y-1}, \qquad
    d t = 0.1 \frac{\min(d x,d y)}{\sqrt{gH}} .
\end{align}

In Fig.~\ref{fig:SWE_Figure}a,b, RC rapidly loses predictive accuracy, with the NRMSE increasing from the beginning of the rollout. 
This is consistent with Table~\ref{tab:SWE_comparison}, where RC reaches the error threshold after only \(6\) forecasting steps. 
VolterraRC improves early-stage accuracy and achieves the lowest NRMSE over the first ten forecasting steps, but its prediction error subsequently increases and reaches the error threshold at step \(26\).
KARC maintains accurate predictions for a longer horizon, extending the VPT to \(40\) steps.
The quantitative results in Table~\ref{tab:SWE_comparison} show that KARC has a slightly higher NRMSE than VolterraRC over the first ten forecasting steps, but achieves a longer VPT and the lowest total time among the compared methods.

Because source and sink terms are excluded in this setup, the system approximately satisfies mass conservation. 
We therefore evaluate physical consistency using the relative mass error in Fig.~\ref{fig:SWE_Figure}c.
KARC exhibits the slowest increase in the relative mass error, whereas RC and VolterraRC accumulate larger deviations over time. 
These results suggest that KARC not only extends the forecasting horizon but also better preserves the mass-conservation structure of the SWE dynamics.

\subsubsection{Text-to-Image Generation}

As an exploratory extension beyond physical dynamical-system forecasting, we further evaluate whether KARC can serve as a lightweight feature forecaster for diffusion sampling acceleration. 
This experiment is motivated by Spectrum~\cite{Han2026Adaptive}, a recent diffusion-acceleration method that forecasts intermediate denoising features by approximating their sampling trajectories with Chebyshev polynomials and fitting the corresponding coefficients by ridge regression.
This formulation is closely aligned with KARC, since Spectrum can be viewed as a Chebyshev instance of the KARC framework: both methods use fixed basis-function expansions followed by closed-form readout training. 
To examine whether alternative basis dictionaries provide comparable acceleration behavior, we adapt KARC to the text-to-image diffusion setting by replacing the Chebyshev feature forecast with Fourier and B-spline basis expansions.

\begin{figure}[!htbp]
    \centering
    \fitfigure{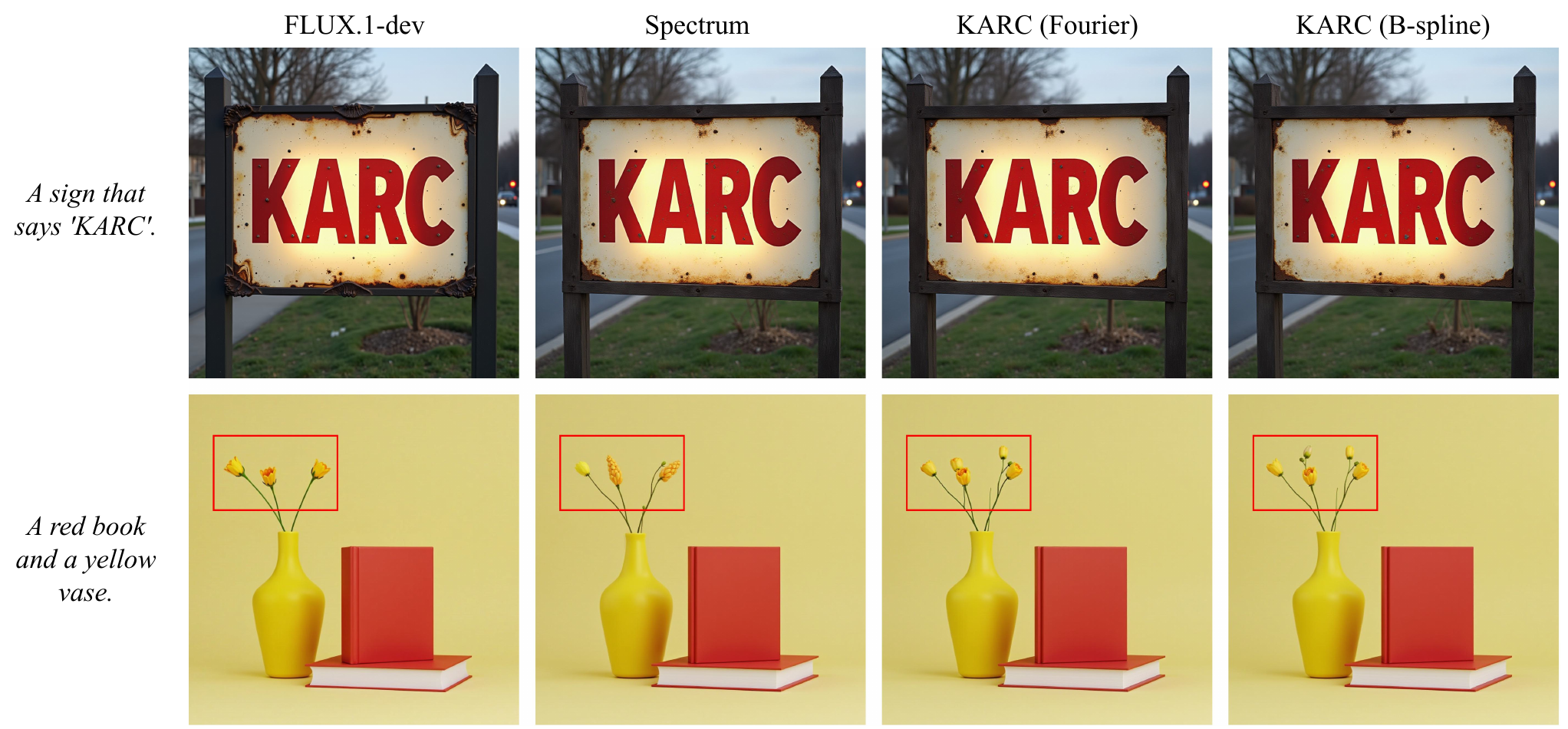}
    \caption{\textbf{Text-to-image generation by Spectrum-based and KARC-based methods.} 
    We integrate KARC into the diffusion sampling process of FLUX.1-dev~\cite{Flux}, as a lightweight feature-forecasting module, replacing the Chebyshev basis function in Spectrum~\cite{Han2026Adaptive} by KARC with Fourier or B-spline functions.
    Each row corresponds to a text prompt. Each column shows the results produced by FLUX.1-dev (baseline), Spectrum, KARC with Fourier bases, and KARC with B-spline bases, respectively. Despite visual differences in the red box, the accelerated methods achieve comparable quantitative performance to the baseline on this task (Table~\ref{tab:Basis_result}).
    }
    \label{fig:Spectrum_Basis}
\end{figure}

\begin{table}[!htbp]
    \centering
    \renewcommand{\arraystretch}{1.25}
    \setlength{\tabcolsep}{6pt}
    \begin{tabular}{lcccccccc}
        \toprule
        \multirow{2}{*}{Method}
        & \multicolumn{4}{c}{\(\alpha\) = 0.75}
        & \multicolumn{4}{c}{\(\alpha\) = 3.0} \\
        \cmidrule(lr){2-5} \cmidrule(lr){6-9}
        & PSNR $\uparrow$ & SSIM $\uparrow$ & LPIPS $\downarrow$ & Speedup $\uparrow$
        & PSNR $\uparrow$ & SSIM $\uparrow$ & LPIPS $\downarrow$ & Speedup $\uparrow$ \\
        \midrule
        Spectrum & 25.058 & 0.868 & 0.123 & 3.391 & 22.358 & 0.803 & 0.203 & 4.665 \\
        KARC (Fourier) & 24.265 & 0.857 & 0.136 & 3.371 & 21.899 & 0.800 & 0.207 & 4.646 \\
        KARC (B-spline) & 24.547 & 0.860 & 0.134 & 3.356 & 22.272 & 0.804 & 0.206 & 4.564 \\
        \bottomrule
    \end{tabular}
    \caption{\textbf{Text-to-image acceleration on the FLUX model evaluated on DrawBench200 for various methods.} 
    Spectrum~\cite{Han2026Adaptive}, KARC (Fourier) and KARC (B-spline) achieve comparable quantitative performance. Here, \(\alpha\) is the scheduling parameter used to control the feature forecasting strategy; see~\cite{Han2026Adaptive} for more details.
    PSNR, SSIM, and LPIPS denote Peak Signal-to-Noise Ratio, Structural Similarity Index Measure, and Learned Perceptual Image Patch Similarity, respectively, and are computed with respect to the original images generated by FLUX.1-dev.
    LPIPS is computed using the AlexNet backbone.}
    \label{tab:Basis_result}
\end{table}

In this experiment, we accelerate FLUX.1-dev~\cite{Flux} by using KARC as a lightweight feature-forecasting module, while keeping all experimental settings consistent with~\cite{Han2026Adaptive}. 
Fig.~\ref{fig:Spectrum_Basis} provides a qualitative comparison of the generated images. 
As shown in the first row, the images generated by accelerating FLUX.1-dev with KARC using Fourier and B-spline bases are visually close to both the original FLUX.1-dev outputs and the Spectrum-accelerated results, preserving the main semantic content and overall image structure. 
The second row further shows that different basis choices can lead to local detail variations, especially in fine-grained object structures, as highlighted by the red boxes. 
These qualitative observations are consistent with the quantitative results in Table~\ref{tab:Basis_result}, where KARC with Fourier and B-spline bases achieves image quality comparable to Spectrum in terms of PSNR, SSIM, and LPIPS, while maintaining similar acceleration performance.

Overall, this experiment suggests that KARC can serve as a lightweight feature forecaster for diffusion sampling acceleration.
The results show that KARC can exploit its explicit basis-function representation to forecast feature trajectories in modern neural network models. 
They also indicate that the KARC formulation is flexible with respect to the choice of basis dictionary, since Fourier and B-spline bases both provide performance close to the Chebyshev-based Spectrum method.
This flexibility supports the broader interpretation of KARC as a general basis-expansion framework rather than a model tied to a single predefined dictionary. 
Therefore, KARC offers a simple and efficient route for extending the basis-function feature forecaster to generative diffusion models.

\section{Discussion}

We have introduced KARC, a data-driven forecasting framework that constructs nonlinear features by expanding delay-embedded coordinates over prescribed univariate basis-function dictionaries.
By training only a linear readout via ridge regression, KARC inherits the closed‑form efficiency of RC while circumventing the sequential dependencies and reducing dependence on hyperparameter tuning.
From a modeling perspective, KARC can be viewed as a delay-coordinate feature-regression framework with fixed univariate basis-function dictionaries, rather than as a conventional reservoir system with internal dynamical states.
We further compare KARC with random nonlinear feature maps, showing that its performance gains arise not merely from high-dimensional nonlinear features, but from the structured features induced by basis-function expansions (see the Supplementary Information~IV).

Although KARC and KANs~\cite{KAN,KAN2} are closely related in mathematical form, KARC is a special realization of the KAN formulation rather than a fully equivalent architecture. 
Specifically, KARC fixes the univariate basis dictionary and uses a linear readout for closed-form training, whereas KANs learn flexible nonlinear functions and compositional transformations through optimization. 
This design substantially improves training efficiency, but may also reduce the expressive capacity of full KANs. 
Higher-order KARC partially compensates for this limitation by introducing coordinate interactions through products of basis responses; however, these interactions are encoded as explicit basis-product features for a linear readout, rather than as fully learnable hierarchical compositions.
Therefore, KARC trades part of the flexibility of KANs for the computational efficiency and simplicity. Developing more expressive KARC variants while retaining closed-form training remains an important future direction.

Moreover, the KAN perspective provides a useful lens for reinterpreting NG-RC. 
NG-RC can be interpreted as a special case of high-order KARC under linear inner functions and polynomial outer interactions (see the Supplementary Information~II).
High-order KARC generalizes this formulation by replacing the linear inner functions with richer univariate basis expansions, before applying polynomial outer interactions. 
Under this interpretation, NG-RC becomes a special case of high-order KARC, while KARC provides a broader KAN-inspired framework for constructing nonlinear autoregressive predictors. 
This interpretation helps explain why high-order KARC can provide richer nonlinear features than NG-RC, although this improvement comes at the cost of a larger feature dimension.

This KAN-inspired view also clarifies the relation between KARC and another recent attempt to combine RC with KANs, i.e., RCKAN~\cite{li2025reservoir}. It explored the integration of RC and KANs by replacing the linear readout of conventional RC with a trainable KAN module.
Although this design can potentially enhance the expressive power of the output mapping by introducing a nonlinear KAN readout, it sacrifices the closed-form training advantage of conventional RC. 
Moreover, RCKAN mainly combines RC and KANs at the architectural level, without explicitly establishing a mathematical connection between the RC readout and the Kolmogorov-Arnold representation. 
In contrast, KARC reformulates the Kolmogorov-Arnold representation as a basis-function feature expansion followed by a linear readout, thereby preserving the closed-form ridge-regression training paradigm of RC while leveraging the representational benefits of basis functions.

We also note the difference between KARC and FNO~\cite{FNO}.
Although both methods may involve Fourier basis representations, they follow different modeling paradigms.
FNO is a neural operator that learns mappings between function spaces by applying trainable spectral kernels in the Fourier domain, thereby capturing global spatial interactions in distributed fields.
In contrast, KARC does not perform an integral transform over the physical spatial domain or learn a spectral operator. 
Instead, it applies coordinate-wise basis expansions to delay-embedded states and represents spatial or temporal coupling through explicit regression features, including high-order products of basis responses. 
Therefore, KARC is better viewed as a feature-regression framework based on delay-coordinate embeddings, whereas FNO is a deep operator-learning framework for modeling mappings between spatial functions.

This distinction suggests that KARC and FNO may be complementary rather than competing approaches. 
In the Supplementary Information~VII.B, we further explored a hybrid KARC-FNO strategy for Navier-Stokes forecasting, where KARC first produces a coarse-scale prediction of the dominant large-scale flow structures and FNO then refines the result by recovering small-scale filamentary vortices.
This preliminary hybrid design suggests a possible route to leveraging the efficient closed-form forecasting capability of KARC to improve computational efficiency without sacrificing predictive accuracy.

More broadly, this work mainly considers deterministic systems, without stochastic perturbations or measurement noise. 
Noise is common in real-world physical processes and may degrade forecasting accuracy or even induce qualitative changes in the dynamics.
Future work will extend KARC to stochastic systems driven by noise~\cite{lin2024learning,lin2025multi}. In addition, the fixed basis dictionary used in KARC may limit its adaptability to systems with abrupt transitions, stiffness, or non-smooth structures. 
Thus, it is interesting to develop adaptive basis-learning strategies such as spline-knot optimization by alternating least squares, and incorporate physical constraints~\cite{Raissi2019Physics,Li2024Physics} to guide structured feature selection in high-dimensional PDE systems. 
These directions may improve the robustness, adaptability, and interpretability of KARC while preserving its efficient closed-form training paradigm.

\section{Methods}

\subsection{High-order Kolmogorov-Arnold Reservoir Computing}

We introduce high-order KARC as an extension of the first-order KARC formulation established in the Results section. 
First-order KARC can be written as a linear-readout approximation of the Kolmogorov-Arnold representation, where the approximation takes the form
\begin{equation}
    f(x_1,\ldots,x_n)
    \approx
    \sum_{p=1}^{n}
    \sum_{j=1}^{m}
    w_{p,j}\psi_j(x_p),
\end{equation}
which consists only of additive contributions from individual delayed coordinates. 
Although this construction preserves a univariate basis-function representation and leads to linear feature scaling, it does not explicitly capture interactions among different coordinates.
To increase nonlinear representational capacity, we replace the linear outer functions with polynomial functions, yielding explicit multiplicative interactions among univariate basis responses.

For high-order KARC, the construction of the inner univariate functions remains the same as in the first-order case. 
That is, each inner function \(\phi_{q,p}\) is still approximated by a finite expansion over the prescribed basis functions:
\begin{equation}
    \phi_{q,p}(x_p)
    \approx
    \sum_{j=1}^{m}
    c_{q,p,j}\psi_j(x_p).
\end{equation}
The key difference is that the outer function is no longer restricted to be linear. 
Instead, we approximate it by a polynomial function up to order \(R\):
\begin{equation}
    \Phi_q(z)
    =
    a_{q,1}z
    +
    a_{q,2}z^2
    +
    \cdots
    +
    a_{q,R}z^R .
\end{equation}
Therefore, the high-order KARC approximation can be written as
\begin{align}
    f(\mathbf{x})
    &\approx
    \sum_{q=1}^{2n+1}
    \Phi_q\left(z_q(\mathbf{x})\right) \\
    &\approx
    \sum_{q=1}^{2n+1}
    \sum_{r=1}^{R}
    a_{q,r}
    \left(
        \sum_{p=1}^{n}
        \sum_{j=1}^{m}
        c_{q,p,j}\psi_j(x_p)
    \right)^r .
\end{align}
For notational simplicity, we define the first-order basis response as
\begin{equation}
    \eta_{p,j}(\mathbf{x})
    =
    \psi_j(x_p),
    \qquad
    p=1,\ldots,n,\quad j=1,\ldots,m .
\end{equation}
Then the high-order approximation can be rewritten as
\begin{align}
    f(\mathbf{x})
    &\approx
    \sum_{q=1}^{2n+1}
    \sum_{r=1}^{R}
    a_{q,r}
    \left(
        \sum_{p=1}^{n}
        \sum_{j=1}^{m}
        c_{q,p,j}\eta_{p,j}(\mathbf{x})
    \right)^r .
\end{align}
Expanding the \(r\)-th power gives
\begin{align}
    f(\mathbf{x})
    &\approx
    \sum_{r=1}^{R}
    \sum_{p_1=1}^{n}
    \sum_{j_1=1}^{m}
    \cdots
    \sum_{p_r=1}^{n}
    \sum_{j_r=1}^{m}
    \left(
        \sum_{q=1}^{2n+1}
        a_{q,r}
        \prod_{\ell=1}^{r}
        c_{q,p_\ell,j_\ell}
    \right)
    \prod_{\ell=1}^{r}
    \eta_{p_\ell,j_\ell}(\mathbf{x}) .
\end{align}
By absorbing the coefficients associated with the inner and outer functions into a single readout coefficient,
\begin{equation}
    w^{(r)}_{p_1,j_1,\ldots,p_r,j_r}
    =
    \sum_{q=1}^{2n+1}
    a_{q,r}
    \prod_{\ell=1}^{r}
    c_{q,p_\ell,j_\ell},
\end{equation}
we obtain the simplified high-order KARC form:
\begin{equation}
    f(\mathbf{x})
    \approx
    \sum_{r=1}^{R}
    \sum_{p_1=1}^{n}
    \sum_{j_1=1}^{m}
    \cdots
    \sum_{p_r=1}^{n}
    \sum_{j_r=1}^{m}
    w^{(r)}_{p_1,j_1,\ldots,p_r,j_r}
    \prod_{\ell=1}^{r}
    \psi_{j_\ell}(x_{p_\ell}).
\end{equation}
This expression shows that high-order KARC remains linear in the trainable readout coefficients, while the nonlinearity is encoded in the high-order products of univariate basis responses. 
Therefore, the model can be written compactly as
\begin{equation}
    \widehat{f}(\mathbf{x})
    =
    \mathbf{W}_{\mathrm{out}}
    \boldsymbol{\Psi}_{\leq R}(\mathbf{x}),
\end{equation}
where \(\boldsymbol{\Psi}_{\leq R}(\mathbf{x})\) denotes the non-redundant basis-product feature vector up to order \(R\), excluding the zeroth-order constant feature, with the total feature dimension
\begin{equation}
    D_R = \sum_{r=1}^{R} \binom{nm+r-1}{r}.
\end{equation}
In practice, we augment \(\boldsymbol{\Psi}_{\leq R}(\mathbf{x})\) with a constant feature to account for the bias term, so that the implemented feature vector has dimension \(D_R+1\).
Thus, high-order KARC preserves the closed-form ridge-regression training structure, while increasing the nonlinear representational capacity through polynomial outer functions.

\subsection{Univariate Basis Dictionary}

We next specify the univariate basis-function dictionaries used to construct the KARC feature map, while using the same time-delay embedding and ridge-regression readout formulation.
The choice of basis determines the inductive bias of the feature representation: Fourier bases
emphasize periodic and oscillatory structures, B-splines capture localized variations, and Chebyshev polynomials provide stable global approximation on bounded domains. 
This modular design separates the construction of nonlinear features from the optimization of the readout, allowing the basis family to be selected according to the structure of the target dynamics without changing the downstream training strategy. 
Below, we summarize the three basis families used in this work.

\textbf{Fourier basis.}
Fourier functions provide a global harmonic dictionary and are well suited for systems with dominant oscillatory or periodic components. Given a period parameter \(P\), we define
\begin{equation}
    \psi_{2i-1}(x)
    =
    \cos\!\left(2\pi\frac{i}{P}x\right),
    \quad
    \psi_{2i}(x)
    =
    \sin\!\left(2\pi\frac{i}{P}x\right),
    \quad
    i=1,2,\ldots,Q.
\end{equation}
The number of Fourier basis functions is therefore \(m=2Q\).
This representation captures smooth periodic or quasi-periodic dynamics using a compact set of frequency components.

\textbf{B-spline basis.}
B-splines provide locally supported basis functions and are therefore suited to localized or nonuniform variations.
Let \(\{a_i\}\) denote the knot sequence. The zeroth-degree basis is defined as
\begin{equation}
    B_{i,0}(x)=
    \begin{cases}
    1,& a_i\le x<a_{i+1},\\
    0,& \text{otherwise},
    \end{cases}
\end{equation}
and higher-degree bases are generated by the Cox-de Boor recursion:
\begin{equation}
    B_{i,s}(x)
    =
    \frac{x-a_i}{a_{i+s}-a_i}B_{i,s-1}(x)
    +
    \frac{a_{i+s+1}-x}{a_{i+s+1}-a_{i+1}}B_{i+1,s-1}(x),
\end{equation}
where \(s\) denotes the spline degree as a hyperparameter and \(\psi_i(x)=B_{i,s}(x)\). 
This local parameterization is useful when the dynamics contain nonuniform regimes, localized structures, or sharp temporal and spatial transitions.

\textbf{Chebyshev basis.}
Chebyshev polynomials form a stable global basis on bounded domains and are widely used in spectral numerical methods. 
They are defined recursively as
\begin{equation}
    T_0(x)=1,\qquad
    T_1(x)=x,\qquad
    T_{n+1}(x)=2xT_n(x)-T_{n-1}(x),
\end{equation}
with \(\psi_i(x)=T_i(x)\). 
Compared with standard monomial polynomial bases, Chebyshev bases generally provide better numerical conditioning and help reduce spurious oscillations in high-order approximations. 
Therefore, they are well suited for smooth but non-periodic dynamics on bounded domains.

In summary, the univariate basis dictionary provides a flexible mechanism for adapting KARC features to different types of dynamical behavior without modifying the readout-training procedure. 
Fourier bases are appropriate when the dominant patterns are oscillatory, B-splines are useful for localized or nonuniform structures, and Chebyshev bases provide a stable global representation on bounded domains. 
Because all three choices are incorporated through the same feature-construction pipeline, the basis family can be treated as a modular modeling component. 
This modularity allows KARC to balance representation quality, numerical stability, and computational efficiency according to the target system.

\subsection{Memory-Optimized Readout Training in High-dimensional Systems}

In high-dimensional systems, estimating the linear readout $\mathbf{W}_{\mathrm{out}}$ of KARC by ridge regression can encounter substantial memory bottlenecks.
Let 
$\mathbf{H}\in\mathbb{R}^{d_h\times N}$ denote the feature matrix,
$\mathbf{Y}\in\mathbb{R}^{d_u \times N}$ denote the target matrix, and 
$\mathbf{W}_{\mathrm{out}}\in\mathbb{R}^{d_u\times d_h}$ denote the readout matrix, where $d_u$ is the system-state dimension, $d_h$ is the KARC feature dimension, and $N$ is the number of training samples. 
The standard ridge-regression solution is
\begin{equation}
    \mathbf{W}_{\mathrm{out}}
    =
    \mathbf{Y}\mathbf{H}^{\top}
    \left(\mathbf{H}\mathbf{H}^{\top}+\lambda \mathbf{I} \right)^{-1}.
\end{equation}
Although this closed-form solution is computationally attractive, directly applying it can become memory-prohibitive when $d_u$, $d_h$, or $N$ is large. 
The bottleneck arises not only from storing $\mathbf{W}_{\mathrm{out}}$, but also from materializing the feature matrix $\mathbf{H}$ and the Gram matrix $\mathbf{H}\mathbf{H}^\top$. 
Our objective is to preserve the exact ridge-regression solution while avoiding large dense matrix materialization.
We adopt three complementary strategies: the Woodbury identity, chunk-wise computation, and low-rank readout factorization.

\textbf{Woodbury identity.}
The first strategy targets the inverse term
$(\mathbf{H}\mathbf{H}^\top+\lambda \mathbf{I})^{-1}\in\mathbb{R}^{d_h\times d_h}$,
which becomes expensive to form and invert when the feature dimension $d_h$ is large. Using the Woodbury identity, we rewrite
\begin{equation}
    \mathbf{H}^\top(\mathbf{H}\mathbf{H}^\top+\lambda \mathbf{I})^{-1}
    =
    (\mathbf{H}^\top\mathbf{H}+\lambda \mathbf{I})^{-1}\mathbf{H}^\top .
\end{equation}
Thus, the readout can be computed as
\begin{equation}
\mathbf{W}_{\mathrm{out}}=
\begin{cases}
\mathbf{Y}\mathbf{H}^\top(\mathbf{H}\mathbf{H}^\top+\lambda \mathbf{I})^{-1}, & d_h<N,\\
\mathbf{Y}(\mathbf{H}^\top\mathbf{H}+\lambda \mathbf{I})^{-1}\mathbf{H}^\top, & d_h\ge N.
\end{cases}
\end{equation}
In the high-dimensional regime, this replaces inversion of a $d_h\times d_h$ matrix with inversion of an $N\times N$ matrix, substantially reducing memory usage when $d_h\gg N$.

\textbf{Chunk-wise computation.}
The second strategy avoids storing the full feature matrix $\mathbf{H}\in\mathbb{R}^{d_h\times N}$ at once. 
We partition the feature dimension into $c$ blocks,
\begin{equation}
    \mathbf{H}=
    \begin{bmatrix}
    \mathbf{H}_1\\
    \mathbf{H}_2\\
    \vdots\\
    \mathbf{H}_c
    \end{bmatrix},
    \qquad
    \mathbf{H}_i\in\mathbb{R}^{d_c\times N},
    \qquad
    d_h=cd_c .
\end{equation}
In the high-dimensional regime $d_h\ge N$, the sample-space Gram matrix can be accumulated block by block as
\begin{equation}
    \mathbf{H}^\top\mathbf{H}
    =
    \sum_{i=1}^{c}\mathbf{H}_i^\top\mathbf{H}_i .
\end{equation}
We first form
\begin{equation}
    \mathbf{G} =
    \sum_{i=1}^{c}\mathbf{H}_i^\top\mathbf{H}_i + \lambda \mathbf{I},
\end{equation}
where $\mathbf{G} \in\mathbb{R}^{N\times N}$, and then compute the readout by feature blocks:
\begin{equation}
    \mathbf{W}_{\mathrm{out}}
    =
    \begin{bmatrix}
    \mathbf{W}_1 & \mathbf{W}_2 & \cdots & \mathbf{W}_c
    \end{bmatrix},
    \qquad
    \mathbf{W}_i=\mathbf{Y} \mathbf{G}^{-1}\mathbf{H}_i^\top,
    \quad i=1,\ldots,c .
\end{equation}
This procedure computes both the Gram matrix and the readout without materializing the full feature matrix, thereby reducing peak memory usage during training.

\textbf{Low-rank readout factorization.}
The third strategy targets the storage of the readout matrix itself. When both the system dimension $d_u$ and the feature dimension $d_h$ are large, storing
$\mathbf{W}_{\mathrm{out}}\in\mathbb{R}^{d_u\times d_h}$ can dominate memory consumption. We therefore approximate the readout by a low-rank factorization:
\begin{equation}
    \mathbf{W}_{\mathrm{out}}\approx \mathbf{A}\mathbf{B},
    \qquad
    \mathbf{A}\in\mathbb{R}^{d_u\times d_l},
    \quad
    \mathbf{B}\in\mathbb{R}^{d_l\times d_h},
\end{equation}
where $d_l\ll \min(d_u,d_h)$. 
The two factors are optimized by alternating least squares:
\begin{align}
    \mathbf{A}^{(k+1)}
    &=
    \mathbf{Y}\mathbf{H}^\top (\mathbf{B}^{(k)})^\top
    \left(
    \mathbf{B}^{(k)}\mathbf{H}\mathbf{H}^\top (\mathbf{B}^{(k)})^\top
    + \lambda \mathbf{I}
    \right)^{-1},
    \notag\\
    \mathbf{B}^{(k+1)}
    &=
    \left(
    (\mathbf{A}^{(k+1)})^\top \mathbf{A}^{(k+1)} + \lambda \mathbf{I}
    \right)^{-1}
    (\mathbf{A}^{(k+1)})^\top \mathbf{Y} \mathbf{H}^\top
    \left(
    \mathbf{H}\mathbf{H}^\top + \lambda \mathbf{I}
    \right)^{-1}.
    \label{eq:als_updates}
\end{align}
Unlike the Woodbury and chunk-wise formulations, which preserve the exact ridge-regression solution, low-rank factorization introduces a controlled approximation to reduce readout storage. 

Together, these strategies make KARC readout training feasible in high-dimensional systems by reducing the memory cost of feature storage, matrix inversion, and readout parameterization. In the experiments reported in this work, we use the Woodbury identity and chunk-wise computation to preserve the exact ridge-regression solution while avoiding full feature-matrix materialization. When the readout matrix \(\mathbf{W}_{\mathrm{out}}\) itself becomes too large to store, we further consider low-rank readout factorization as an additional approximation strategy. The impact of this approximation on forecasting performance is evaluated in Supplementary Information~III.

\subsection{Error Bound of KARC for Fading Memory Dynamical Systems}
We provide a one-step error analysis for KARC applied to fading-memory dynamical systems, which offers a qualitative basis for understanding how hyperparameters affect model performance.
Specifically, we consider a class of discrete-time systems whose next state depends on the past trajectory, and assume that the corresponding history-to-state map satisfies both the fading memory property and Lipschitz continuity with respect to a weighted norm.
Before presenting the error bound, we first introduce the dynamical setting and the required assumptions.

Given a discrete-time dynamical system with memory, we write its evolution as
\begin{equation}
    \mathbf{u}_{i+1}=F(\mathbf{u}_{i}^{-}),
\end{equation}
where \(\mathbf{u}_{i}\in\mathbb{R}^{d}\), and
\begin{equation}
    \mathbf{u}_{i}^{-}
    =
    (\ldots,\mathbf{u}_{i-2},\mathbf{u}_{i-1},\mathbf{u}_{i})
    \in(\mathbb{R}^{d_u})^{\mathbb{Z}_{-}}
\end{equation}
denotes the left-infinite history of the system up to time \(i\).
In practice, we restrict the admissible histories to a uniformly bounded set whose elements take values in \([0,1]^d_u\). Specifically, we define
\begin{equation}
    K = ([0,1]^{d_u})^{\mathbb{Z}_{-}} .
\end{equation}
Then the history-to-state map is given by
\begin{equation}
    F:K\rightarrow \mathbb{R}^{d_u},  
\end{equation}
which determines the next state from the entire past trajectory.

We next introduce the weighted norm used to quantify the influence of the past trajectory.
Let \(w:\mathbb{N}_{0}\to(0,1]\) be a decreasing weighting sequence satisfying
\begin{equation}
    \lim_{k\to\infty} w_k=0 .   
\end{equation}
For any left-infinite history \(\mathbf{z} \in K\), the weighted norm associated with \(w\) is defined as
\begin{equation}
    \lVert \mathbf{z} \rVert_{w}
    :=
    \sup_{t\in\mathbb{Z}_{-}}
    \left \{
    \lVert \mathbf{z}_{t} \rVert w_{-t}
    \right \}.
\end{equation}
Equivalently, for two histories
\(
    \mathbf{u}_{i}^{-}
    =
    (\ldots,\mathbf{u}_{i-2},\mathbf{u}_{i-1},\mathbf{u}_{i}),
\)
and
\(
    \mathbf{v}_{i}^{-}
    =
    (\ldots,\mathbf{v}_{i-2},\mathbf{v}_{i-1},\mathbf{v}_{i}),
\)
their weighted distance is given by
\begin{equation}
    \lVert \mathbf{u}_{i}^{-}-\mathbf{v}_{i}^{-} \rVert_{w}
    :=
    \sup_{k\ge 0}
    \left \{
    w_{k}
    \lVert \mathbf{u}_{i-k}-\mathbf{v}_{i-k} \rVert
    \right \}.
\end{equation}
Since \(w_{k}\to 0\), discrepancies in the remote past are assigned progressively smaller weights.

Under this weighted norm, the history-dependent system is assumed to satisfy the fading memory property.
Specifically, the map \(F\) satisfies the fading memory property if there exists a weighting sequence \(w\) such that
\begin{equation}
    F:(K,\lVert\cdot\rVert_{w})\to\mathbb{R}^{d_u}
\end{equation}
is continuous.
That is, for any \(\mathbf{u}^{-}\in K\) and any \(\varepsilon>0\), there exists \(\delta(\varepsilon)>0\) such that, for any \(\mathbf{v}^{-}\in K\),
\begin{equation}
    \lVert \mathbf{u}^{-}-\mathbf{v}^{-} \rVert_{w}
    <
    \delta(\varepsilon)
    \quad
    \Longrightarrow
    \quad
    \lVert F(\mathbf{u}^{-})-F(\mathbf{v}^{-}) \rVert
    <
    \varepsilon .
\end{equation}
This means that the next state depends continuously on the past trajectory, while the influence of perturbations in the remote past fades according to the weighting sequence \(w\).

We further assume that \(F\) is Lipschitz continuous with respect to the weighted norm \(\lVert\cdot\rVert_{w}\).
Specifically, there exists a constant \(L_{F}>0\) such that, for any two histories \(\mathbf{u}^{-},\mathbf{v}^{-}\in K\),
\begin{equation}
    \lVert F(\mathbf{u}^{-})-F(\mathbf{v}^{-}) \rVert
    \le
    L_{F} \lVert \mathbf{u}^{-}-\mathbf{v}^{-} \rVert_{w}.
\end{equation}
This condition provides a quantitative form of fading memory: the change in the next state is bounded by the weighted distance between two histories.
Since Lipschitz continuity implies continuity, this assumption is stronger than the fading memory property.

Since the true map \(F\) is defined on the left-infinite history space, it is generally impossible to approximate \(F(\mathbf{u}^{-})\) by using all past states in practical computation. 
Therefore, we first introduce a finite time-delay truncation of the history. 
For a delay length \(k\), define
\begin{equation}
    \mathbf{u}_{i}^{-k}
    =
    [\mathbf{u}_{i-k+1}^{\top},\ldots,\mathbf{u}_{i-1}^{\top},\mathbf{u}_{i}^{\top}]^{\top}
    \in\mathbb{R}^{kd_u},
\end{equation}
which is equivalent to the delay-embedded vector \(\mathbf{x}_i\) introduced in Eq.(\ref{equ:delayed_embedding}) (i,e.,\(\mathbf{u}_{i}^{-k} \equiv \mathbf{x}_i\)).
Instead of directly learning the infinite-history map \(F(\mathbf{u}_{i}^{-})\), KARC learns a finite-delay surrogate map
\begin{equation}
    \widehat{F}_{k}(\mathbf{u}_{i}^{-k})
    \approx
    F(\mathbf{u}_{i}^{-}).
\end{equation}
The KARC predictor is then written as
\begin{equation}
    \widehat{\mathbf{u}}_{i+1}
    =
    \mathbf{W}_{\mathrm{out}}
    \mathbf{\Psi}(\mathbf{u}_{i}^{-k}),
\end{equation}
where \(\mathbf{\Psi}(\cdot)\) denotes the KARC feature map and \(\mathbf{W}_{\mathrm{out}}\) is the learned linear readout.
Therefore, for a given history \(\mathbf{u}_{i}^{-}\), the total one-step prediction error can be written as
\begin{equation}
\begin{aligned}
    e_{\mathrm{tot}}
    &=
    \left\lVert
    F(\mathbf{u}_{i}^{-})
    -
    \mathbf{W}_{\mathrm{out}}
    \mathbf{\Psi}(\mathbf{u}_{i}^{-k})
    \right\rVert _2 \\
    &=
    \left\lVert
    \left(
    F(\mathbf{u}_{i}^{-})
    -
    \widehat{F}_{k}(\mathbf{u}_{i}^{-k})
    \right)
    +
    \left(
    \widehat{F}_{k}(\mathbf{u}_{i}^{-k})
    -
    \mathbf{W}_{\mathrm{out}}
    \mathbf{\Psi}(\mathbf{u}_{i}^{-k})
    \right)
    \right\rVert_2.
\end{aligned}
\end{equation}
The first term corresponds to the time-delay truncation error, which measures the loss caused by replacing the infinite history \(\mathbf{u}_{i}^{-}\) with the finite delay vector \(\mathbf{u}_{i}^{-k}\). 
The second term corresponds to the KARC approximation error for the finite-delay surrogate system. We assume that the finite-delay surrogate map can be decomposed as
\begin{equation}
    \widehat{F}_{k}(\mathbf{u}_{i}^{-k})
    =
    \widehat{\mathbf{W}}
    \mathbf{\Psi}(\mathbf{u}_{i}^{-k})
    +
    \mathbf{r}(\mathbf{u}_{i}^{-k}),
\end{equation}
where \(\widehat{\mathbf{W}}\) denotes the ideal readout matrix within the chosen KARC feature space, and \(\mathbf{r}(\mathbf{u}_{i}^{-k})\) is the residual term caused by the finite expressiveness of the feature map. 
Substituting this decomposition into the total error gives
\begin{equation}
\begin{aligned}
    e_{\mathrm{tot}}
    &=
    \left\lVert
    \left(
    F(\mathbf{u}_{i}^{-})
    -
    \widehat{F}_{k}(\mathbf{u}_{i}^{-k})
    \right)
    +
    \mathbf{r}(\mathbf{u}_{i}^{-k})
    +
    \left(
    \widehat{\mathbf{W}}
    -
    \mathbf{W}_{\mathrm{out}}
    \right)
    \mathbf{\Psi}(\mathbf{u}_{i}^{-k})
    \right\rVert_2 .
\end{aligned}
\end{equation}

Let the training feature matrix and target matrix be defined as
\begin{equation}
    \mathbf{H}
    =
    \left[
    \mathbf{\Psi}(\mathbf{u}_{1}^{-k}),
    \mathbf{\Psi}(\mathbf{u}_{2}^{-k}),
    \ldots,
    \mathbf{\Psi}(\mathbf{u}_{N}^{-k})
    \right],
\end{equation}
and
\begin{equation}
    \mathbf{Y}
    =
    \left[
    \mathbf{u}_{2},
    \mathbf{u}_{3},
    \ldots,
    \mathbf{u}_{N+1}
    \right].
\end{equation}
The ridge regression solution for the KARC readout is
\begin{equation}
    \mathbf{W}_{\mathrm{out}}
    =
    \mathbf{Y}\mathbf{H}^{\top}
    \left(
    \mathbf{H}\mathbf{H}^{\top}
    +
    \lambda \mathbf{I}
    \right)^{-1},
\end{equation}
where \(\lambda >0\) is the ridge regularization parameter.
 For the finite-delay surrogate system, we write the training targets as
\begin{equation}
    \mathbf{Y}
    =
    \widehat{\mathbf{W}}\mathbf{H}
    +
    \mathbf{R},
\end{equation}
where
\begin{equation}
    \mathbf{R}
    =
    \left[
    \mathbf{r}(\mathbf{u}_{1}^{-k}),
    \mathbf{r}(\mathbf{u}_{2}^{-k}),
    \ldots,
    \mathbf{r}(\mathbf{u}_{N}^{-k})
    \right]
\end{equation}
collects the residual terms on the training samples. For simplicity, define
\[ 
    \mathbf{G} :=
    \mathbf{H}\mathbf{H}^{\top}
    +
    \lambda \mathbf{I},
    \quad 
    \mathbf{\Psi}_i^{-k}: = \mathbf{\Psi}(\mathbf{u}_{i}^{-k}).
\]
Then the learned readout can be rewritten as
\begin{equation}
\begin{aligned}
    \mathbf{W}_{\mathrm{out}}
    &=
    \mathbf{Y}\mathbf{H}^{\top}\mathbf{G}^{-1} \\
    &=
    \left(
    \widehat{\mathbf{W}}\mathbf{H}
    +
    \mathbf{R}
    \right)
    \mathbf{H}^{\top}
    \mathbf{G}^{-1} \\
    &=
    \widehat{\mathbf{W}}
    \mathbf{H}\mathbf{H}^{\top}
    \mathbf{G}^{-1}
    +
    \mathbf{R}\mathbf{H}^{\top}\mathbf{G}^{-1}.
\end{aligned}
\end{equation}
Since
\begin{equation}
    \mathbf{H}\mathbf{H}^{\top}
    =
    \mathbf{G}
    -
    \lambda\mathbf{I},
\end{equation}
we have
\begin{equation}
\begin{aligned}
    \mathbf{W}_{\mathrm{out}}
    &=
    \widehat{\mathbf{W}}
    \left(
    \mathbf{G}
    -
    \lambda\mathbf{I}
    \right)
    \mathbf{G}^{-1}
    +
    \mathbf{R}\mathbf{H}^{\top}\mathbf{G}^{-1} \\
    &=
    \widehat{\mathbf{W}}
    -
    \lambda
    \widehat{\mathbf{W}}
    \mathbf{G}^{-1}
    +
    \mathbf{R}\mathbf{H}^{\top}\mathbf{G}^{-1}.
\end{aligned}
\end{equation}
Therefore,
\begin{equation}
    \widehat{\mathbf{W}}
    -
    \mathbf{W}_{\mathrm{out}}
    =
    \lambda
    \widehat{\mathbf{W}}
    \mathbf{G}^{-1}
    -
    \mathbf{R}\mathbf{H}^{\top}\mathbf{G}^{-1}.
\end{equation}
Substituting this expression back into the error decomposition gives
\begin{equation}
\begin{aligned}
    e_{\mathrm{tot}}
    &=
    \left\lVert
    \left(
    F(\mathbf{u}_{i}^{-})
    -
    \widehat{F}_{k}(\mathbf{u}_{i}^{-k})
    \right)
    +
    \mathbf{r}(\mathbf{u}_{i}^{-k})
    +
    \left(
    \widehat{\mathbf{W}}
    -
    \mathbf{W}_{\mathrm{out}}
    \right)
    \mathbf{\Psi}_i^{-k}
    \right\rVert_2 \\
    &=
    \left\lVert
    \left(
    F(\mathbf{u}_{i}^{-})
    -
    \widehat{F}_{k}(\mathbf{u}_{i}^{-k})
    \right)
    +
    \mathbf{r}(\mathbf{u}_{i}^{-k})
    +
    \lambda \widehat{\mathbf{W}}\mathbf{G}^{-1}\mathbf{\Psi}_i^{-k}
    -
    \mathbf{R}\mathbf{H}^{\top}\mathbf{G}^{-1}\mathbf{\Psi}_i^{-k}
    \right\rVert_2 .
\end{aligned}
\end{equation}
Using the triangle inequality, the total error can be decomposed into four terms:
\begin{equation}
    \begin{aligned}
        e_{\mathrm{tot}}
        &\le
        \left\lVert
        F(\mathbf{u}_{i}^{-})
        -
        \widehat{F}_{k}(\mathbf{u}_{i}^{-k})
        \right\rVert_2
        +
        \left\lVert
        \mathbf{r}(\mathbf{u}_{i}^{-k})
        \right\rVert_2  
        +
        \lambda
        \left\lVert
        \widehat{\mathbf{W}}
        \mathbf{G}^{-1}
        \mathbf{\Psi}_i^{-k}
        \right\rVert_2
        +
        \left\lVert
        \mathbf{R}
        \mathbf{H}^{\top}
        \mathbf{G}^{-1}
        \mathbf{\Psi}_i^{-k}
        \right\rVert_2 .
    \end{aligned}
\end{equation}

The first term corresponds to the time-delay truncation error caused by replacing the infinite history \(\mathbf{u}_{i}^{-}\) with the finite delay vector \(\mathbf{u}_{i}^{-k}\). 
To compare these two objects in the weighted history space, we lift the finite-delay vector \(\mathbf{u}_{i}^{-k}\) to a left-infinite sequence by padding the discarded remote past with a fixed reference value, here chosen as zero:
\begin{equation}
    \widetilde{\mathbf{u}}_{i}^{-k}
    =
    (\ldots,\mathbf{0},\mathbf{0},
    \mathbf{u}_{i-k+1},\ldots,\mathbf{u}_{i-1},\mathbf{u}_{i}).
\end{equation}
Then the finite-delay surrogate system can be written as
\begin{equation}
    \widehat{F}_{k}(\mathbf{u}_{i}^{-k})
    =
    F(\widetilde{\mathbf{u}}_{i}^{-k}).
\end{equation}
Therefore, by the Lipschitz continuity of \(F\) with respect to the weighted norm, we have
\begin{equation}
\begin{aligned}
    e_{\mathrm{delay}}
    &=
    \left\lVert
    F(\mathbf{u}_{i}^{-})
    -
    \widehat{F}_{k}(\mathbf{u}_{i}^{-k})
    \right\rVert_{2} \\
    &=
    \left\lVert
    F(\mathbf{u}_{i}^{-})
    -
    F(\widetilde{\mathbf{u}}_{i}^{-k})
    \right\rVert_{2} \\
    &\le
    L_{F}
    \left\lVert
    \mathbf{u}_{i}^{-}
    -
    \widetilde{\mathbf{u}}_{i}^{-k}
    \right\rVert_{w}.
\end{aligned}
\end{equation}
Since \(\mathbf{u}_{i}^{-}\) and \(\widetilde{\mathbf{u}}_{i}^{-k}\) share the most recent \(k\) states, their difference only comes from the discarded remote past. Hence,
\begin{equation}
\begin{aligned}
    \left\lVert
    \mathbf{u}_{i}^{-}
    -
    \widetilde{\mathbf{u}}_{i}^{-k}
    \right\rVert_{w}
    &=
    \sup_{\ell\ge k}
    \left\{
    w_{\ell}
    \left\lVert
    \mathbf{u}_{i-\ell}
    \right\rVert_{2}
    \right\}.
\end{aligned}
\end{equation}
Since \(\mathbf{u}_{i}^{-}\in K\), each state in the history belongs to \([0,1]^d_u\). Therefore,
\begin{equation}
    \left\lVert
    \mathbf{u}_{t}
    \right\rVert_{2}
    \le
    \sqrt{d_u},
    \qquad
    \forall t .
\end{equation}
Then, since \(w_{\ell}\) is decreasing, we have
\begin{equation}
    \begin{aligned}
        \left\lVert
        \mathbf{u}_{i}^{-}
        -
        \widetilde{\mathbf{u}}_{i}^{-k}
        \right\rVert_{w}
        &\le
        \sqrt{d_u}
        \sup_{\ell\ge k} w_{\ell}  \\
        &=
        \sqrt{d_u} w_{k}.
    \end{aligned}
\end{equation}
Therefore,
\begin{equation}
    e_{\mathrm{delay}}
    \le
    L_{F} \sqrt{d_u} w_{k}.
\end{equation}

The second term represents the approximation error from the finite KARC feature dictionary. We assume that
\begin{equation}
\left\lVert
\mathbf{r}(\mathbf{u}_{i}^{-k})
\right\rVert_2
\le
\varepsilon_{\Psi}(k),
\end{equation}
where \(\varepsilon_{\Psi}(k)\) denotes the residual approximation error of the finite KARC feature dictionary for the \(k\)-delay target map. It measures how well the truncated history-to-state map can be represented in the feature space spanned by \(\mathbf{\Psi}\). This quantity depends on the delay length \(k\), the chosen basis family, and the number of basis functions. A richer feature dictionary generally reduces \(\varepsilon_{\Psi}(k)\), while a limited dictionary leads to a larger residual approximation error.

For the third term, we use the submultiplicativity of the spectral norm to obtain
\begin{equation}
    \begin{aligned}
        \lambda
        \left\lVert
        \widehat{\mathbf{W}}
        \mathbf{G}^{-1}
        \mathbf{\Psi}_i^{-k}
        \right\rVert_2
        &\le
        \lambda
        \left\lVert
        \widehat{\mathbf{W}}
        \right\rVert_2
        \left\lVert
        \mathbf{G}^{-1}
        \right\rVert_2
        \left\lVert
        \mathbf{\Psi}_i^{-k}
        \right\rVert_2 .
    \end{aligned}
\end{equation}
We assume that the ideal readout matrix and the KARC feature vector are uniformly bounded, namely,
\begin{equation}
    \left\lVert
    \widehat{\mathbf{W}}
    \right\rVert_2
    \le
    B_W,
    \qquad
    \left\lVert
    \mathbf{\Psi}_i^{-k}
    \right\rVert_2
    \le
    B_{\Psi},
\end{equation}
where \(B_W\) characterizes the coefficient norm of the ideal finite-dimensional KARC approximation, and \(B_{\Psi}\) bounds the magnitude of the feature representation on the admissible input domain. Thus, the third term is bounded as
\begin{equation}
    \lambda
    \left\lVert
    \widehat{\mathbf{W}}
    \mathbf{G}^{-1}
    \mathbf{\Psi}_i^{-k}
    \right\rVert_2
    \le
    \lambda B_W B_{\Psi}
    \left\lVert
    \mathbf{G}^{-1}
    \right\rVert_2 .
\end{equation}

Similarly, the fourth term can be controlled by the submultiplicativity of the spectral norm:
\begin{equation}
    \begin{aligned}
        \left\lVert
        \mathbf{R}
        \mathbf{H}^{\top}
        \mathbf{G}^{-1}
        \mathbf{\Psi}_i^{-k}
        \right\rVert_2
        &\le
        \left\lVert
        \mathbf{R}\mathbf{H}^{\top}
        \right\rVert_2
        \left\lVert
        \mathbf{G}^{-1}
        \right\rVert_2
        \left\lVert
        \mathbf{\Psi}_i^{-k}
        \right\rVert_2 .
    \end{aligned}
\end{equation}
We assume that the residuals and the training feature vectors are uniformly bounded, i.e.,
\begin{equation}
    \left\lVert
    \mathbf{r}(\mathbf{u}_{j}^{-k})
    \right\rVert_2
    \le
    \varepsilon_{\Psi}(k),
    \qquad
    \left\lVert
    \mathbf{\Psi}_{j}^{-k}
    \right\rVert_2
    \le
    B_{\Psi},
    \qquad
    j=1,\ldots,N .
\end{equation}
Since
\begin{equation}
    \mathbf{R}\mathbf{H}^{\top}
    =
    \sum_{j=1}^{N}
    \mathbf{r}(\mathbf{u}_{j}^{-k})
    \left(\mathbf{\Psi}_{j}^{-k}\right)^{\top},
\end{equation}
we have
\begin{equation}
    \begin{aligned}
        \left\lVert
        \mathbf{R}\mathbf{H}^{\top}
        \right\rVert_2
        &\le
        \sum_{j=1}^{N}
        \left\lVert
        \mathbf{r}(\mathbf{u}_{j}^{-k})
        \left(\mathbf{\Psi}_{j}^{-k}\right)^{\top}
        \right\rVert_2  \\
        &\le
        \sum_{j=1}^{N}
        \left\lVert
        \mathbf{r}(\mathbf{u}_{j}^{-k})
        \right\rVert_2
        \left\lVert
        \mathbf{\Psi}_{j}^{-k}
        \right\rVert_2  \\
        &\le
        N B_{\Psi}\varepsilon_{\Psi}(k),
    \end{aligned}
\end{equation}
where \(N\) denotes the number of training samples. Combining this bound with
\(\left\lVert \mathbf{\Psi}_i^{-k}\right\rVert_2 \le B_{\Psi}\), we obtain
\begin{equation}
    \left\lVert
    \mathbf{R}
    \mathbf{H}^{\top}
    \mathbf{G}^{-1}
    \mathbf{\Psi}_i^{-k}
    \right\rVert_2
    \le
    N B_{\Psi}^{2}
    \varepsilon_{\Psi}(k)
    \left\lVert
    \mathbf{G}^{-1}
    \right\rVert_2 .
\end{equation}

To make the bound explicit, recall that
\(\mathbf{G}=\mathbf{H}\mathbf{H}^{\top}+\lambda I\).
Since the eigenvalues of \(\mathbf{H}\mathbf{H}^{\top}\) are the squared singular values of \(\mathbf{H}\), we have
\begin{equation}
    \left\lVert
    \mathbf{G}^{-1}
    \right\rVert_2
    =
    \frac{1}
    {\sigma_{\min}(\mathbf{H})^2+\lambda},
\end{equation}
where \(\sigma_{\min}(\mathbf{H})\) denotes the smallest singular value of \(\mathbf{H}\).
To further specify the feature bound \(B_{\Psi}\), let \(n=kd_u\) denote the dimension of the delay-embedded input and let \(m\) be the number of univariate basis functions used for each input coordinate. For the first-order KARC feature map
\begin{equation}
\mathbf{\Psi}(\mathbf{x})
=
\big[
\psi_1(x_1),\ldots,\psi_m(x_1),
\ldots,
\psi_1(x_n),\ldots,\psi_m(x_n)
\big]^{\top},
\end{equation}
the constant \(B_{\Psi}\) can be chosen according to the boundedness of the selected basis family.

\textbf{Fourier Basis.}
We use a constant basis together with paired sine and cosine functions. 
Specifically, for each input coordinate, the Fourier dictionary has the form
\begin{equation}
    \{\cos(\omega_1 x),\sin(\omega_1 x),\ldots,
    \cos(\omega_Q x),\sin(\omega_Q x)\},
\end{equation}
where \(m=2Q\). Since $\sin^2(\omega_q x)+\cos^2(\omega_q x)=1$, the squared norm of the Fourier feature vector for each coordinate satisfies
\begin{equation}
    \sum_{q=1}^{Q}
    \left[
    \cos^2(\omega_q x)+\sin^2(\omega_q x)
    \right]
    =
    Q
    =
    \frac{m}{2}.
\end{equation}
Therefore, for the delay-embedded input \(\mathbf{x}\in\mathbb{R}^{n}\), we have
\begin{equation}
    \left\lVert
    \mathbf{\Psi}(\mathbf{x})
    \right\rVert_2
    \le
    B_{\Psi}^{\mathrm{Fourier}}
    =
    \sqrt{\frac{nm}{2}}.
\end{equation}

\textbf{B-spline Basis.}
For standard normalized B-spline bases on the knot-covered domain, the basis functions are nonnegative and form a partition of unity:
\begin{equation}
    0 \le B_{j,s}(x) \le 1,
    \qquad
    \sum_{j=1}^{m} B_{j,s}(x)=1 .
\end{equation}
Therefore, the squared norm of the B-spline feature vector for each coordinate can be bounded by
\begin{equation}
    \sum_{j=1}^{m} B_{j,s}(x)^2
    \le
    \left(
    \sum_{j=1}^{m} B_{j,s}(x)
    \right)^2
    =
    1 .
\end{equation}
Under the standard partition-of-unity normalization, each coordinate contributes at most one to the squared feature norm, and hence the concatenated B-spline feature vector over \(n\) delay coordinates satisfies
\begin{equation}
    \left\lVert
    \mathbf{\Psi}(\mathbf{x})
    \right\rVert_2
    \le
    B_{\Psi}^{\mathrm{B\text{-}spline}}
    =
    \sqrt{n}.
\end{equation}

\textbf{Chebyshev Basis.}
For Chebyshev basis, the admissible input domain is \(x\in[0,1]\). Since \([0,1]\subset[-1,1]\), the Chebyshev polynomials of the first kind satisfy
\begin{equation}
|T_j(x)| = |\cos(j\arccos x)| \le 1,
\qquad
x\in[0,1].
\end{equation}
Therefore, each Chebyshev basis response is uniformly bounded in magnitude by one. Since the first-order KARC feature vector concatenates \(m\) Chebyshev basis responses for each of the \(n\) delay coordinates, we obtain
\begin{equation}
    \left\lVert
    \mathbf{\Psi}(\mathbf{x})
    \right\rVert_2
    \le
    B_{\Psi}^{\mathrm{Chebyshev}}
    =
    \sqrt{nm}.
\end{equation}

Combining the above basis-dependent bounds, the feature bound \(B_{\Psi}\) can be chosen as
\begin{equation}
    B_{\Psi}
    =
    \begin{cases}
        \sqrt{\dfrac{nm}{2}},
        & \text{Fourier basis}\\
        \sqrt{n},
        & \text{B-spline basis}\\
        \sqrt{nm},
        & \text{Chebyshev basis}.
    \end{cases}
\end{equation}
Substituting the bound on \(\left\lVert \mathbf{G}^{-1}\right\rVert_2\) into the previous error decomposition, we obtain the final one-step error bound
\begin{equation}
\label{ErrorBound}
    \begin{aligned}
        e_{\mathrm{tot}}
        &\le
        L_{F} \sqrt{d_u} w_{k}
        +
        \varepsilon_{\Psi}(k)
        \left(
        1
        +
        \frac{N B_{\Psi}^{2}}
        {\sigma_{\min}(\mathbf{H})^2+\lambda}
        \right)
        +
        \frac{\lambda B_W B_{\Psi}}
        {\sigma_{\min}(\mathbf{H})^2+\lambda}.
    \end{aligned}
\end{equation}
This bound shows that the one-step prediction error of KARC consists of three main components. The first term, \(L_{F} \sqrt{d_u} w_{k}\), is the time-delay truncation error caused by replacing the infinite history with a finite delay vector. The second term, involving \(\varepsilon_{\Psi}(k)\), is the approximation error induced by the finite KARC feature dictionary, together with its propagation through ridge regression. The last term is the ridge regularization bias, which is controlled by the regularization parameter \(\lambda\), the ideal readout norm bound \(B_W\), and the basis-dependent feature bound \(B_{\Psi}\).

This decomposition further shows that the effect of hyperparameters is not necessarily monotonic. 
For instance, increasing the delay length \(k\) can reduce the time-delay truncation error \(L_F\sqrt{d_u}w_k\), but it also enlarges the delay-embedded input space and may affect both the finite-dictionary approximation error \(\varepsilon_{\Psi}(k)\) and the conditioning of the feature matrix through \(\sigma_{\min}(\mathbf{H})\).
Therefore, a larger delay length does not necessarily lead to a smaller overall error. 
Similar trade-offs also arise from the choice of basis-function type, the number of basis functions, and the regularization parameter \(\lambda\).
Thus, this bound mainly provides a qualitative guide for understanding different error sources, while practical hyperparameter selection is performed empirically.

\textbf{Data availability:}
The authors declare that the data supporting this study are available within the paper.

\textbf{Code availability:}
The present algorithm will be publicly available upon the acceptance of the manuscript.

\section*{Acknowledgments}
\addtocontents{toc}{\protect\setcounter{tocdepth}{0}}
\addtocontents{toc}{\protect\setcounter{tocdepth}{3}}
We thank Lu Zhong and Xiaonan Gao for their helpful communication. This work is supported by Project 12322501, 12575035 of the National Natural Science Foundation of China, and 2026NSFSCZY0124 of the Natural Science Foundation of Sichuan Province. 
The computational work is supported by the Center for HPC, University of Electronic Science and Technology of China. 

\section*{Author contributions}
\addtocontents{toc}{\protect\setcounter{tocdepth}{0}}
\addtocontents{toc}{\protect\setcounter{tocdepth}{3}}
Y.T. and J.H. had the original idea for this work. J.H. performed the study with the guidance from Y.T. and J.K., and all authors contributed to the preparation of the manuscript.

\section*{Competing interests}
\addtocontents{toc}{\protect\setcounter{tocdepth}{0}}
\addtocontents{toc}{\protect\setcounter{tocdepth}{3}}

The authors declare no competing interests.

\section*{Additional information}
\addtocontents{toc}{\protect\setcounter{tocdepth}{0}}
\addtocontents{toc}{\protect\setcounter{tocdepth}{3}}

\textbf{Supplementary information} The online version
contains supplementary material available at [URL will be inserted by publisher].

\textbf{Correspondence and requests for materials} should be addressed to Ying Tang.

\textbf{Reprints and permission information}  is available online at  [URL will be inserted by publisher].

\bibliography{bib}
















\end{document}


\title{Supplementary Information: Kolmogorov-Arnold Reservoir Computing}

\author{Juntian Huang}
\affiliation{Institute of Fundamental and Frontier Sciences, University of Electronic Science and Technology of China, Chengdu 611731, China}

\author{Jürgen Kurths}
\affiliation{Potsdam Institute for Climate Impact Research, Potsdam 14412, Germany}
\affiliation{Department of Physics, Humboldt University Berlin, Berlin 12489, Germany}
\affiliation{\mbox{Research Institute of Intelligent Complex Systems, Fudan University, Shanghai 200433, China}}

\author{Ying Tang}
\email[Corresponding authors: ]{jamestang23@gmail.com}
\affiliation{Institute of Fundamental and Frontier Sciences, University of Electronic Science and Technology of China, Chengdu 611731, China}
\affiliation{School of Physics, University of Electronic Science and Technology of China, Chengdu 611731, China}
\affiliation{Key Laboratory of Quantum Physics and Photonic Quantum Information, Ministry of Education, University of Electronic Science and Technology of China, Chengdu 611731, China}
\affiliation{Non-classical Information Science Basic Discipline Research Center of Sichuan Province, University of Electronic Science and Technology of China, Chengdu 611731, China}

\maketitle
\tableofcontents
\clearpage

\section{Baseline formulations for comparison}

This section briefly reviews the baseline reservoir-computing formulations used for comparison in the main text and in the subsequent hyperparameter sensitivity analysis.
We consider three representative approaches: conventional reservoir computing (RC), which relies on a fixed recurrent reservoir; next generation reservoir computing (NG-RC), which replaces recurrence with explicit polynomial features on delay coordinates~\cite{NG-RC}; and VolterraRC, which extends NG-RC to an infinite-dimensional kernel formulation~\cite{VolterraRC}. 
For each method, we summarize the core formulation and the training procedure, providing the necessary background for the comparative experiments reported in the main text and the supplementary sensitivity studies below.

\subsection{Reservoir Computing}

RC is a lightweight recurrent neural network framework for time-series modelling and dynamical-system prediction. In contrast to recurrent neural networks, where both recurrent and output-layer parameters are typically optimized through backpropagation, RC projects input signals into a fixed high-dimensional nonlinear dynamical system and trains only a linear readout. Because the reservoir parameters remain unchanged after initialization, RC substantially reduces training complexity while retaining rich temporal processing capability. This property has made RC particularly attractive for chaotic-system forecasting, nonlinear time-series analysis and spatiotemporal dynamical modelling.

Conventional RC is mainly represented by two paradigms: echo state networks (ESNs) \cite{ESN} and liquid state machines (LSMs) \cite{LSM}. ESNs usually employ discrete-time random recurrent networks as reservoirs, whereas LSMs are inspired by spiking neural networks and biological neural dynamics. Despite these architectural differences, both paradigms rely on an untrained internal dynamical system to transform input sequences into high-dimensional transient state representations through fixed nonlinear responses.

Taking the ESN as an example, a commonly used leaky reservoir update is given by
\begin{equation}
    \mathbf{r}_{t+1}=(1-\gamma) \mathbf{r}_t+\gamma f\left(\mathbf{A}\mathbf{r}_t+\mathbf{W}_{\mathrm{in}}x_t+\mathbf{b}\right),
\end{equation}
where \(\mathbf{r}_t\) denotes the reservoir state at time \(t\), \(\mathbf{x}_t\) is the input, \(\mathbf{A}\) is the recurrent reservoir matrix, \(\mathbf{W}_{\mathrm{in}}\) is the input weight matrix, \(\mathbf{b}\) is a bias term, \(\gamma\) is the leakage rate, and \(f(\cdot)\) is a nonlinear activation function. The matrices \(\mathbf{A}\) and \(\mathbf{W}_{\mathrm{in}}\) are typically randomly initialized and then kept fixed during training. Through this recurrent update, the reservoir acts as a nonlinear temporal embedding mechanism: the state \(\mathbf{r}_t\) depends not only on the current input \(\mathbf{x}_t\), but also on the history of previous inputs encoded in the reservoir dynamics.

The prediction is produced by a trainable linear readout,
\begin{equation}
    \mathbf{y}_t=\mathbf{W}_{\mathrm{out}}\mathbf{r}_t,
\end{equation}
where \(\mathbf{W}_{\mathrm{out}}\) is the readout matrix. Training an RC model therefore reduces to estimating \(\mathbf{W}_{\mathrm{out}}\), commonly by least-squares regression or ridge regression. With \(\ell_2\) regularization, the closed-form ridge-regression solution can be written as
\begin{equation}
    \mathbf{W}_{\mathrm{out}}
    =
    \mathbf{Y}\mathbf{H}^{\top}\left(\mathbf{H}\mathbf{H}^{\top}+\lambda \mathbf{I}\right)^{-1},
\end{equation}
where \(\mathbf{H}\) is the matrix of collected reservoir states, \(\mathbf{Y}\) is the corresponding target-output matrix and \(\lambda\) is the regularization coefficient. 
The regularization term improves the conditioning of the readout estimation and helps reduce overfitting.

\subsection{Next Generation Reservoir Computing}

NG-RC~\cite{NG-RC} is an extension of RC designed to simplify the reservoir construction while preserving nonlinear temporal modelling capability.
RC uses a fixed random recurrent reservoir to generate high-dimensional nonlinear state representations through recursive updates. 
By contrast, NG-RC removes the explicit recurrent reservoir and reformulates reservoir computing as a nonlinear vector autoregression model~\cite{NVAR}. 
The key idea is to construct delay-coordinate representations directly from historical observations, enrich them through polynomial feature expansion, and use a linear readout for prediction.

Unlike RC, where temporal information is encoded implicitly in the evolving reservoir state, NG-RC first forms a delay-coordinate vector from recent observations:
\begin{equation}
    \mathbb{O}_{\mathrm{lin},t}
    =
    \mathbf{x}_t \oplus \mathbf{x}_{t-1} \oplus \cdots \oplus \mathbf{x}_{t-k+1},
\end{equation}
where \(\oplus\) denotes vector concatenation, \(k\) is the delay-embedding length, and \(\mathbf{x}_t\) is the observed system state at time \(t\). 
The vector \(\mathbb{O}_{\mathrm{lin},t}\) combines the current observation with its recent history, thereby providing an explicit temporal memory. 
Thus, NG-RC replaces the recurrent memory mechanism of reservoirs with time-delay coordinates that describe the recent evolution of the system.

To recover the nonlinear representation capability typically provided by recurrent reservoir dynamics, NG-RC augments the delay-coordinate vector with explicit polynomial features. 
Let \(\lceil\otimes\rceil\) denote an operator that collects all unique monomials generated by tensor products while removing duplicate terms. 
The \(p\)-th order nonlinear feature vector is defined as
\begin{equation}
    \mathbb{O}^{(p)}_{\mathrm{nonlin},t}
    =
    \underbrace{
    \mathbb{O}_{\mathrm{lin},t}
    \lceil\otimes\rceil
    \mathbb{O}_{\mathrm{lin},t}
    \lceil\otimes\rceil \cdots
    \lceil\otimes\rceil
    \mathbb{O}_{\mathrm{lin},t}
    }_{p \ \mathrm{times}} .
\end{equation}
The full NG-RC feature vector is then obtained by concatenating a constant bias term, the linear delay-coordinate features and the polynomial nonlinear features:
\begin{equation}
    \mathbb{O}_{\mathrm{total},t}
    =
    c \oplus \mathbb{O}_{\mathrm{lin},t}
    \oplus \mathbb{O}^{(2)}_{\mathrm{nonlin},t}
    \oplus \cdots
    \oplus \mathbb{O}^{(p)}_{\mathrm{nonlin},t}.
\end{equation}
In practice, the feature composition can be tailored to the experimental setting by retaining only selected components. For example, one may use the linear delay-coordinate features together with the third-order nonlinear features:
\begin{equation}
    \mathbb{O}_{\mathrm{total},t}
    =
    \mathbb{O}_{\mathrm{lin},t}
    \oplus
    \mathbb{O}^{(3)}_{\mathrm{nonlin},t}.
\end{equation}
And the output is also obtained through a linear readout,
\begin{equation}
    \mathbf{y}_t = \mathbf{W}_{\mathrm{out}} \mathbb{O}_{\mathrm{total},t},
\end{equation}
where \(\mathbf{W}_{\mathrm{out}}\) is learned by ridge regression, following the same training strategy as in RC.

By replacing recursive reservoir-state updates with explicit delay-coordinate and polynomial feature construction, NG-RC enables parallel feature generation across time steps. However, the dimension of \(\mathbb{O}_{\mathrm{total},t}\) grows rapidly with both the input dimension and the polynomial order \(p\). This rapid growth can lead to a pronounced curse of dimensionality, particularly for high-dimensional dynamical systems.

\subsection{VolterraRC}

VolterraRC can be regarded as an infinite-dimensional extension of NG-RC, obtained in the limiting case where the delay length \(k\) and the polynomial order \(p\) tend to infinity~\cite{VolterraRC}.
In standard NG-RC, the readout is constructed from a finite-dimensional polynomial feature vector \(\mathbb{O}_{\mathrm{total},t}\), which consists of the linear delayed observables and their higher-order monomial combinations. 
Equivalently, NG-RC can be kernelized and formulated as a polynomial-kernel ridge regression:
\begin{equation}
    \mathbf{y}_t 
    = \mathbf{W}_{\mathrm{out}} \mathbb{O}_{\mathrm{total},t}
    = \sum_{i=1}^{N} \alpha_i 
    K^{\mathrm{poly}}\left(
    \mathbb{O}_{\mathrm{lin},t},
    \mathbb{O}_{\mathrm{lin},i}
    \right),
\end{equation}
where \(N\) denotes the number of training samples, and the polynomial kernel is defined as
\begin{equation}
    K^{\mathrm{poly}}(\mathbb{O}_{\mathrm{lin},t},\mathbb{O}_{\mathrm{lin},i})
    =
    \left(1+\mathbb{O}_{\mathrm{lin},t}^\top \mathbb{O}_{\mathrm{lin},i}\right)^p .
\end{equation}
Here, \(\mathbb{O}_{\mathrm{lin},t}\) denotes the linear delayed observable vector, while \(\mathbb{O}_{\mathrm{total},t}\) represents the full polynomial feature vector used in NG-RC. 
The constant component of \(\mathbb{O}_{\mathrm{total},t}\) is set to \(1\), corresponding to the bias term in the polynomial kernel. 
This kernelized formulation avoids the explicit construction of all polynomial features and shows that NG-RC can be interpreted as a finite-dimensional polynomial kernel method.

VolterraRC further extends this idea by replacing the finite-order polynomial kernel with the Volterra kernel, which implicitly incorporates infinitely many delays and polynomial orders. Specifically, the Volterra Gram matrix is defined as
\begin{equation}
    \mathbf{K}^{\mathrm{Volt}}
    =
    \left[
    K^{\mathrm{Volt}}_{i,j}
    \right]_{i,j=1}^{N}
    \in \mathbb{R}^{N \times N},
\end{equation}
where each entry is given by
\begin{equation}
    K^{\mathrm{Volt}}_{i,j}
    =
    K^{\mathrm{Volt}}\left(
    \mathbf{x}_i,
    \mathbf{x}_j
    \right),
    \quad i,j \in \{1,\cdots,N\}.
\end{equation}
The entries of the Volterra Gram matrix can be computed recursively as
\begin{equation}
    K_{i,j}^{\mathrm{Volt}}
    =
    1
    +
    \frac{
    \alpha^2 K_{i-1,j-1}^{\mathrm{Volt}}
    }{
    1-\theta^2
    \left\langle
    \mathbf{x}_i,
    \mathbf{x}_j
    \right\rangle
    },
\end{equation}
with the boundary condition
\begin{equation}
    K_{i,j}^{\mathrm{Volt}}
    =
    \frac{1}{1-\theta^2},
    \quad \text{if } i=0 \text{ or } j=0 .
\end{equation}
Here, \(\alpha\) and \(\theta\) are kernel parameters controlling the contribution of past lags and nonlinear interactions, respectively. 
Therefore, VolterraRC provides a richer feature representation for dynamical systems with long-term memory and high-order nonlinear dependencies, while avoiding the explicit construction of an infinite-dimensional feature space.

\section{NG-RC as a specific KAN-inspired feature model}

In this section, we show that NG-RC can also be derived as a specific construction from the Kolmogorov-Arnold representation perspective used to formulate KARC in the main text. Following the notation in the main text, the Kolmogorov-Arnold representation can be written as
\begin{equation}
    f(\mathbf{x})
    =
    \sum_{q=1}^{2n+1}
    \Phi_q
    \left(
    \sum_{p=1}^{n}
    \phi_{q,p}(x_{p})
    \right),
\end{equation}
where \(\mathbf{x}\in\mathbb{R}^{n}\) is the delay-coordinate input, \(\phi_{q,p}\) denotes the inner univariate function applied to the \(p\)-th coordinate, and \(\Phi_q\) denotes the corresponding outer univariate function. Under this formulation, different choices of the inner and outer functions lead to different feature constructions. This provides a useful way to relate NG-RC to the KARC framework.

NG-RC is recovered when the inner functions are chosen as fixed linear coordinate functions and the outer functions are set as polynomial functions. Specifically, let
\begin{equation}
\phi_{q,p}(x_{p}) = a_{q,p} x_{p},  
\end{equation}
and
\begin{equation}
\Phi_q(z)
=
\sum_{r=0}^{R}
c_{q,r} z^r .
\end{equation}
Substituting these choices into the Kolmogorov-Arnold representation gives
\begin{equation}
    f(\mathbf{x})
    =
    \sum_{q=1}^{2n+1}
    \sum_{r=0}^{R}
    c_{q,r}
    \left(
    \sum_{p=1}^{n}
    a_{q,p}x_{p}
    \right)^r .
\end{equation}
Thus, the nonlinear representation is generated not by nonlinear inner functions, but by polynomial outer interactions applied to linear combinations of the delay-coordinate variables.

Expanding the polynomial terms yields
\begin{equation}
    f(\mathbf{x})
    =
    \sum_{q=1}^{2n+1} c_{q,0}
    +
    \sum_{q=1}^{2n+1} c_{q,1}
    \left(
    \sum_{p=1}^{n} a_{q,p}x_p
    \right)
    +
    \sum_{q=1}^{2n+1} c_{q,2}
    \left(
    \sum_{p=1}^{n} a_{q,p}x_p
    \right)^2
    + \cdots
    +
    \sum_{q=1}^{2n+1} c_{q,R}
    \left(
    \sum_{p=1}^{n} a_{q,p}x_p
    \right)^R .
\end{equation}
The first-order term produces linear features of the input coordinates, whereas the higher-order terms generate polynomial interactions such as \(x_i x_j\), \(x_i x_j x_k\), and other cross-coordinate products up to order \(R\). After collecting terms with the same monomial structure, the model can be written as
\begin{equation}
f(\mathbf{x})
    =
    w_0
    +
    \sum_{p=1}^{n} w_p x_p
    +
    \sum_{p_1,p_2=1}^{n} w_{p_1,p_2}x_{p_1}x_{p_2}
    + \cdots
    +
    \sum_{p_1,\ldots,p_R=1}^{n}
    w_{p_1,\ldots,p_R}
    x_{p_1}\cdots x_{p_R}.
\end{equation}
Therefore, the resulting feature space consists of polynomial features of the delay-coordinate vector up to order \(R\), matching the feature construction used in NG-RC:
\begin{equation}
f(\mathbf{x})
    =
    \mathbf{W}_{\mathrm{out}}
    \mathbb{O}_{\mathrm{total},t}.
\end{equation}
Here, \(\mathbb{O}_{\mathrm{total},t}\) contains the linear and higher-order polynomial features of the delay-coordinate vector.

From this viewpoint, NG-RC can be interpreted as a specific KAN-inspired feature construction in which the inner univariate functions are fixed to be linear and nonlinear expressivity is introduced through polynomial outer interactions. 
This interpretation also clarifies the motivation of KARC. 
Instead of relying on raw linear coordinates followed by polynomial expansion, KARC enriches the inner univariate representation using fixed basis dictionaries, such as Fourier, B-spline, or Chebyshev bases. 
The resulting feature construction retains the closed-form ridge-regression training strategy while allowing a richer set of structured nonlinear features.

\clearpage
\section{Rank sensitivity of low-rank readout factorization}
\label{sec:low_rank_readout}

\begin{figure}[!htbp]
    \centering
    \fitfigure{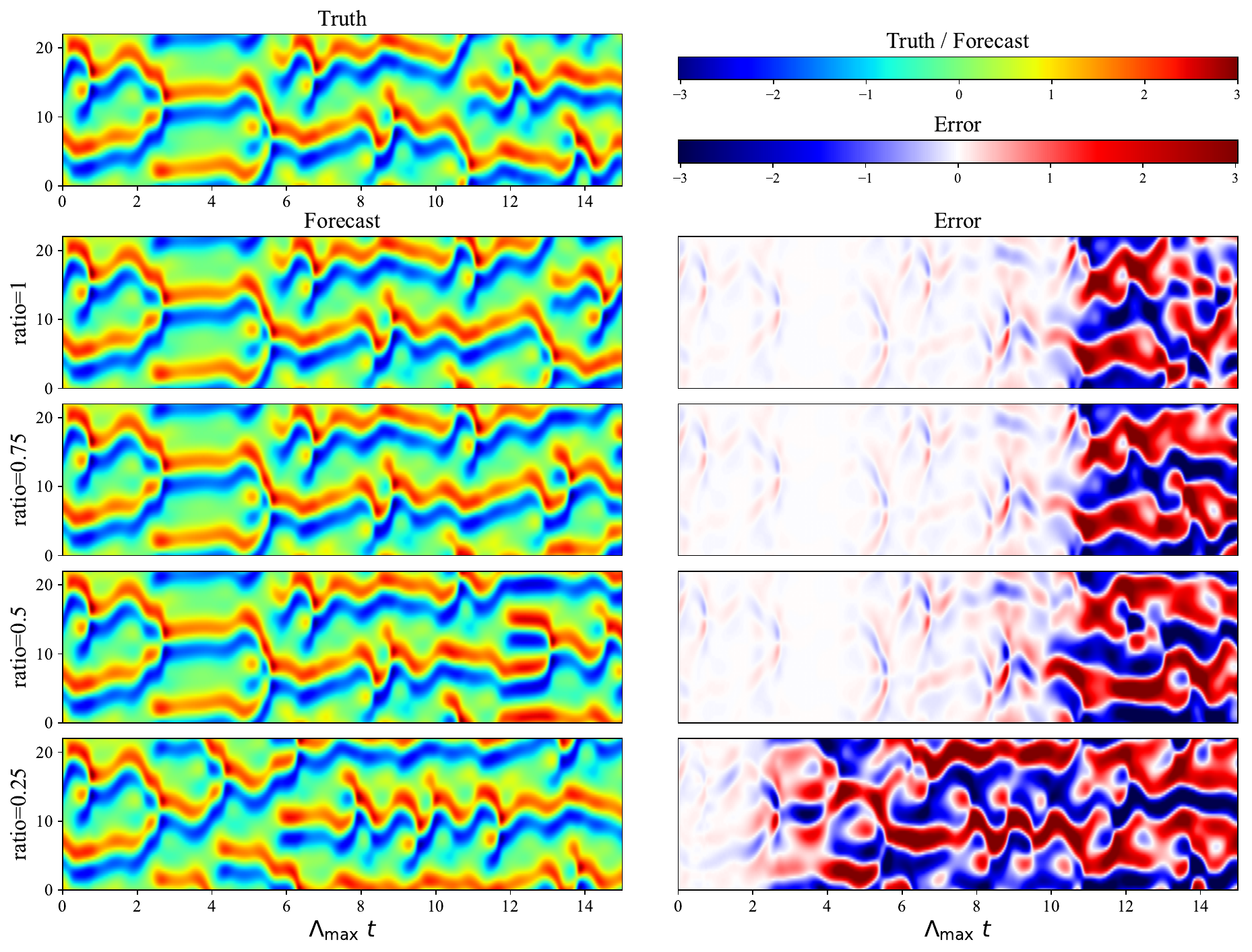}
    \caption{\textbf{Effect of low-rank readout factorization on KARC forecasting for the Kuramoto-Sivashinsky equation.} Forecasting results are shown for the Kuramoto-Sivashinsky equation with domain size L=22, using KARC with Fourier basis functions. The top row shows the ground-truth spatiotemporal field. The following rows show KARC forecasts with different low-rank readout ratios, where ratio=\(d_l / d_h\), \(d_l\) 
	is the low-rank readout dimension and \(d_h\) is the original feature dimension. The left column shows the predicted fields, and the right column shows the corresponding pointwise prediction errors.}
    \label{fig:Low_rank_KS}
\end{figure}

\begin{table}[!htbp]
    \centering
    \setlength{\tabcolsep}{12pt}
    \renewcommand{\arraystretch}{1.25}
    \begin{tabular}{cccc}
        \toprule
         Compression Ratio &  Total Time (s) \(\downarrow\) & NRMSE$@$1LT \(\downarrow\)& VPT ($\epsilon=0.1$) [LT] \(\uparrow\)\\
        \midrule
        $1$  & $ 8.21 $ & $6.300 \times 10^{-2}$ & $10.075$ \\
        $0.75$  & $ 8.15 $ & $ 6.304 \times 10^{-2}$ & $10.100 $ \\ 
        $0.5$ & $8.13 $ & $6.319 \times 10^{-2}$ & $8.312$ \\
        $0.25$ & $8.09 $ & $5.484 \times 10^{-2}$ & $2.062$ \\
        \bottomrule
    \end{tabular}
    \caption{\textbf{Quantitative comparison of KARC with different low-rank readout ratios on the Kuramoto-Sivashinsky equation.}
    The evaluation metrics follow the main-text Kuramoto-Sivashinsky experiments, including total time, NRMSE at one Lyapunov time, and valid prediction time (VPT) with threshold \(\epsilon=0.1\).}
    \label{tab:Lowrank_KS_table}
\end{table}

To evaluate the influence of low-rank readout factorization, we performed experiments on the Kuramoto-Sivashinsky equation with domain size \(L=22\), using KARC with Fourier basis functions.
The compression ratio is defined as \(d_l/d_h\), where \(d_l\) denotes the low-rank readout dimension and \(d_h\) denotes the original feature dimension.
As shown in Figure~\ref{fig:Low_rank_KS} and Table~\ref{tab:Lowrank_KS_table}, moderate rank reduction has a limited influence on short-term accuracy and total time. 
When the ratio is reduced from \(1\) to \(0.75\), the total time slightly decreases from \(8.21\) s to \(8.15\) s, while the NRMSE at one Lyapunov time and the VPT remain almost unchanged.
Further reducing the ratio to \(0.5\) still preserves the dominant spatiotemporal structures, although the VPT decreases to \(8.312\) LT.
These results suggest that, in practical applications, it is not always necessary to store the full readout matrix \(\mathbf{W}_{\mathrm{out}}\); a moderately compressed readout can reduce memory pressure while retaining much of the predictive performance.

However, excessive compression substantially degrades long-term forecasting. When the ratio is reduced to \(0.25\), the short-term NRMSE at one Lyapunov time remains small, \(5.484\times10^{-2}\), but the VPT drops sharply to \(2.062\) LT. 
This indicates that short-term error alone does not fully characterize the autonomous rollout stability of the model. 
The corresponding forecast in Figure~\ref{fig:Low_rank_KS} also shows earlier error growth and visible loss of spatiotemporal coherence. 
These results suggest that moderate low-rank factorization can reduce memory cost with little loss in short-term accuracy, whereas excessive compression compromises long-horizon stability.

\section{Ablation: structured basis features versus random nonlinear features}

\begin{figure}[!htbp]
    \centering
    \fitfigure{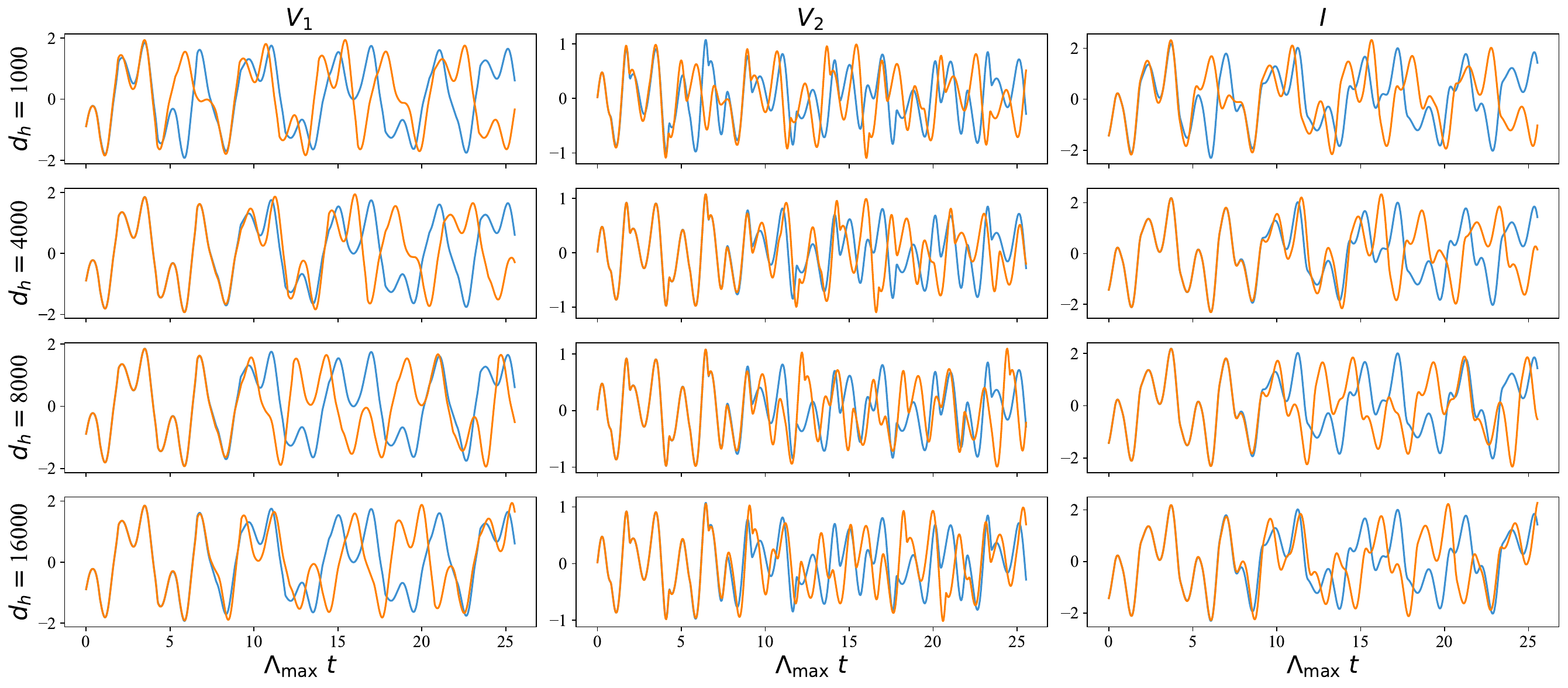}
    \caption{\textbf{Forecasting performance of random MLP features with different feature dimensions on the double-scroll system.}
    Rows correspond to different random-feature dimensions \(d_h\), and columns represent the three state variables \(V_1\), \(V_2\), and \(I\). The random features are generated by a fixed two-layer MLP, and only the final linear readout is trained by ridge regression. Orange and blue curves denote the predicted and reference trajectories, respectively.}
    \label{fig:MLP_DoubleScroll}
\end{figure}

\begin{figure}[!htbp]
    \centering
    \fitfigure{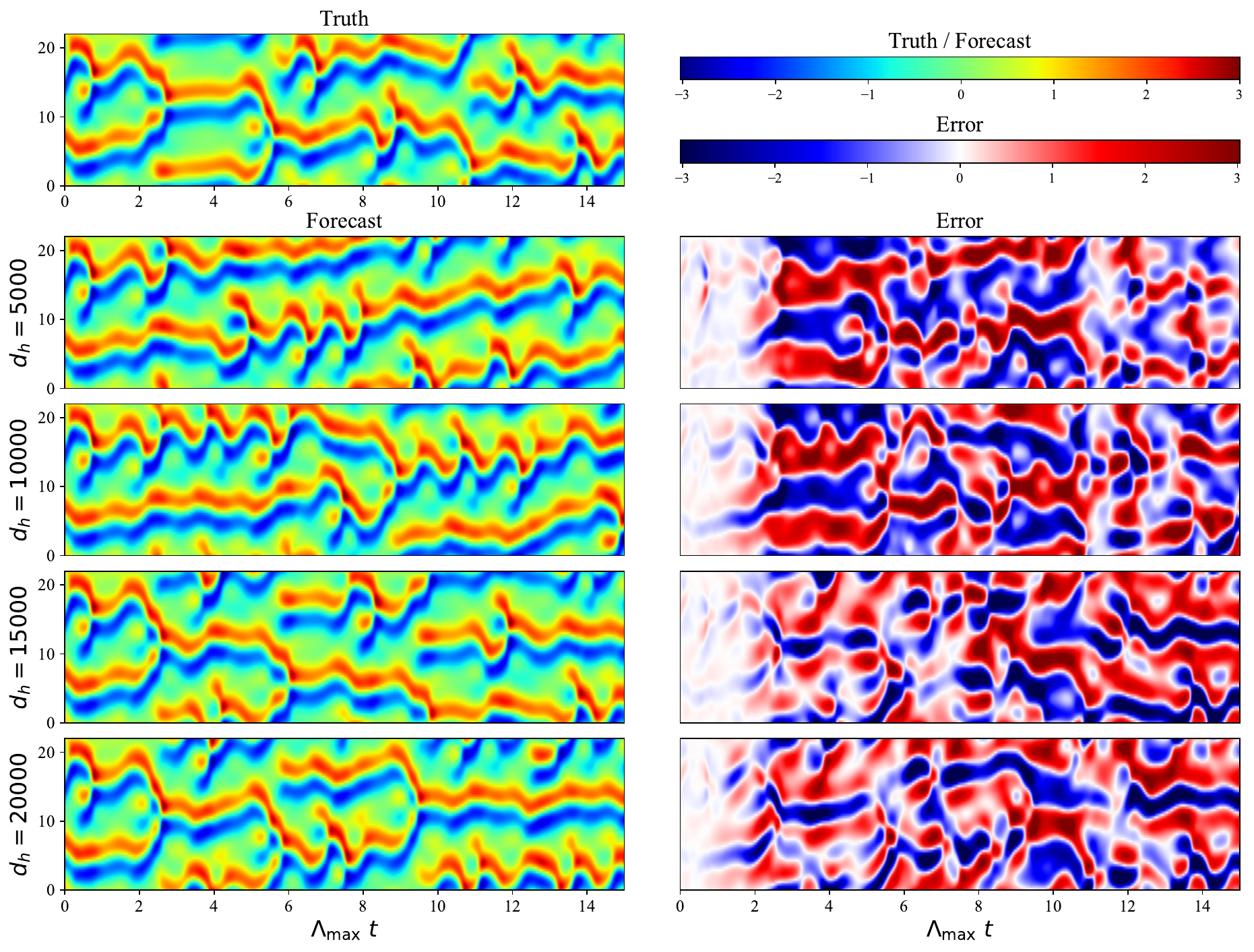}
    \caption{\textbf{Forecasting performance of random MLP features with different feature dimensions on the Kuramoto-Sivashinsky equation.}
    The top row shows the reference spatiotemporal field for the Kuramoto-Sivashinsky equation with domain size \(L=22\). The remaining rows show forecasts generated using random MLP features with different feature dimensions \(d_h\), together with the corresponding pointwise prediction errors.}
    \label{fig:MLP_KS}
\end{figure}

\begin{table}[!htbp]
    \centering
    \setlength{\tabcolsep}{14pt}
    \renewcommand{\arraystretch}{1.25}
    \begin{tabular}{cccc}
        \toprule
         Dimension \(d_h\)  &  Total Time (s) \(\downarrow\) & NRMSE$@$1LT \(\downarrow\)& VPT ($\epsilon=0.1$) [LT] \(\uparrow\)\\
        \midrule
        $1000$  & $0.064$ & $ 5.300 \times 10^{-2}$ & $2.656$ \\
        $4000$  & $0.181$ & $ 1.179 \times 10^{-3}$ & $10.752$ \\ 
        $8000$ & $0.567$ & $9.575 \times 10^{-4}$ & $10.240 $ \\
        $16000$ & $2.098$ & $1.481 \times 10^{-3}$ & $8.480$ \\
        \bottomrule
    \end{tabular}
    \caption{\textbf{Quantitative comparison of random MLP features with different feature dimensions on the double-scroll system.} The evaluation metrics are the same as those used in the previous experiments.}
    \label{tab:MLP_DoubleScroll_table}
\end{table}

\begin{table}[!htbp]
    \centering
    \setlength{\tabcolsep}{14pt}
    \renewcommand{\arraystretch}{1.25}
    \begin{tabular}{cccc}
        \toprule
         Dimension \(d_h\)  &  Total Time (s) \(\downarrow\) & NRMSE$@$1LT \(\downarrow\)& VPT ($\epsilon=0.1$) [LT] \(\uparrow\)\\
        \midrule
        $5000$  & $0.634$ & $ 1.513 \times 10^{-1}$ & $0.625$ \\
        $10000$  & $2.063$ & $ 9.145 \times 10^{-2}$ & $1.075$ \\ 
        $15000$ & $4.632$ & $7.993 \times 10^{-2}$ & $1.425$ \\
        $20000$ & $7.649$ & $8.736 \times 10^{-2}$ & $1.113$ \\
        \bottomrule
    \end{tabular}
    \caption{\textbf{Quantitative comparison of random MLP features with different feature dimensions on the Kuramoto-Sivashinsky equation.} The evaluation metrics are the same as those used in the previous experiments.}
    \label{tab:MLP_KS_table}
\end{table}

To verify the effectiveness of the structured basis features used in KARC, we compare them with random nonlinear features generated by a fixed multilayer perceptron (MLP). 
This comparison serves as a control experiment to distinguish the effect of feature dimensionality from that of feature structure. 
If high-dimensional nonlinear expansion alone were sufficient, random MLP features with comparable or larger dimensions should achieve similar forecasting accuracy after the same linear readout training. 
If KARC remains superior, its gain is more likely to come from the structured univariate basis expansion rather than from dimensional expansion alone. 
Therefore, this experiment examines whether the advantage of KARC arises from the basis structure itself.

In the random-feature baseline, the delay-coordinate input \(\mathbf{x}\) is passed through a randomly initialized two-layer MLP feature map. Specifically, the random nonlinear features are constructed as
\begin{align}
    \mathbf{z}
    &=
    \sigma
    \left(
    \mathbf{W}_1 \mathbf{x}
    +
    \mathbf{b}_1
    \right),\\
    \mathbf{h}
    &=
    \sigma
    \left(
    \mathbf{W}_2 \mathbf{z}
    +
    \mathbf{b}_2
    \right),
\end{align}
where \(\mathbf{W}_1,\mathbf{W}_2\) and \(\mathbf{b}_1,\mathbf{b}_2\) are randomly initialized and kept fixed, and \(\sigma(\cdot)\) denotes the nonlinear activation function. 
In this experiment, the dimension of the first-layer representation \(\mathbf{z}\) is set to be twice that of the final random feature vector \(\mathbf{h}\). 
The prediction is then obtained through a linear readout,
\begin{equation}
    \hat{\mathbf{y}}
    =
    \mathbf{W}_{\mathrm{out}}
    \mathbf{h},
\end{equation}
where only \(\mathbf{W}_{\mathrm{out}}\) is trained. 
Thus, the random-feature baseline follows the same closed-form readout-training strategy as KARC, while replacing the structured basis features with fixed random nonlinear features.

We first evaluate the random MLP feature baseline on the double-scroll system with different feature dimensions.
The results in Figure~\ref{fig:MLP_DoubleScroll} and Table~\ref{tab:MLP_DoubleScroll_table} show that the forecasting performance depends strongly on the random-feature dimension \(d_h\). 
For \(d_h=1000\), the random MLP features provide only a short prediction horizon, with a VPT of \(2.656\) LT. 
Increasing the dimension to \(d_h=4000\) or \(d_h=8000\) substantially improves the results, yielding VPTs of \(10.752\) LT and \(10.240\) LT, respectively.
However, further increasing the dimension to \(d_h=16000\) does not improve the forecast and instead reduces the VPT to \(8.480\) LT, while also increasing the total computational time. 
These results indicate that random MLP features can achieve reasonable forecasting performance on the double-scroll system when the feature dimension is properly chosen, and their performance can be comparable to that of RC.
Nevertheless, the random MLP feature baseline still does not outperform KARC with Fourier basis functions, suggesting that the structured basis expansion used in KARC provides a more effective feature representation than generic random nonlinear projections.

We further test the random MLP feature baseline on the Kuramoto-Sivashinsky equation with domain size \(L=22\). Figure~\ref{fig:MLP_KS} shows that increasing the random-feature dimension does not lead to a substantial improvement in forecasting performance on this more challenging spatiotemporal system.
Across the tested feature dimensions, the forecasts typically deviate from the reference solution after approximately one Lyapunov time, and the accumulated errors grow rapidly during autonomous rollout. 
This suggests that generic random nonlinear projections are insufficient to capture the structured spatiotemporal dependencies of the Kuramoto-Sivashinsky dynamics. In contrast, the main-text results show that KARC with Fourier basis functions uses a feature dimension of \(12801\) and achieves a valid prediction time of approximately \(12\) Lyapunov times. 
This comparison indicates that the advantage of KARC does not simply come from using a high-dimensional nonlinear feature space, but from the structured basis expansion that provides a more suitable representation for chaotic spatiotemporal forecasting.

Overall, the random MLP baseline controls for feature dimensionality while using the same closed-form readout training as KARC. 
Although it performs reasonably on the low-dimensional double-scroll system after dimension tuning, it does not yield a comparable prediction horizon on the Kuramoto–Sivashinsky equation. 
The forecasts typically remain accurate for only about one Lyapunov time, even as the random-feature dimension is increased. 
These results suggest that KARC's advantage arises not from high dimensionality alone, but from structured basis features that are better suited to spatiotemporal chaotic dynamics.

\clearpage

\section{Additional hyperparameter sensitivity analyses}

This section presents a comprehensive hyperparameter sensitivity analysis for all competing methods. Specifically, we vary the reservoir dimension for RC, the polynomial expansion order for NG-RC, the memory kernel parameter \(\alpha\) for VolterraRC, and the basis-function family for KARC. 
These additional experiments serve two purposes: first, to justify the default parameter choices adopted in the main text; and second, to illustrate the performance variability of each method under different configurations.

\subsection{Double-Scroll system}
\begin{figure}[!htbp]
    \centering
    \fitfigure{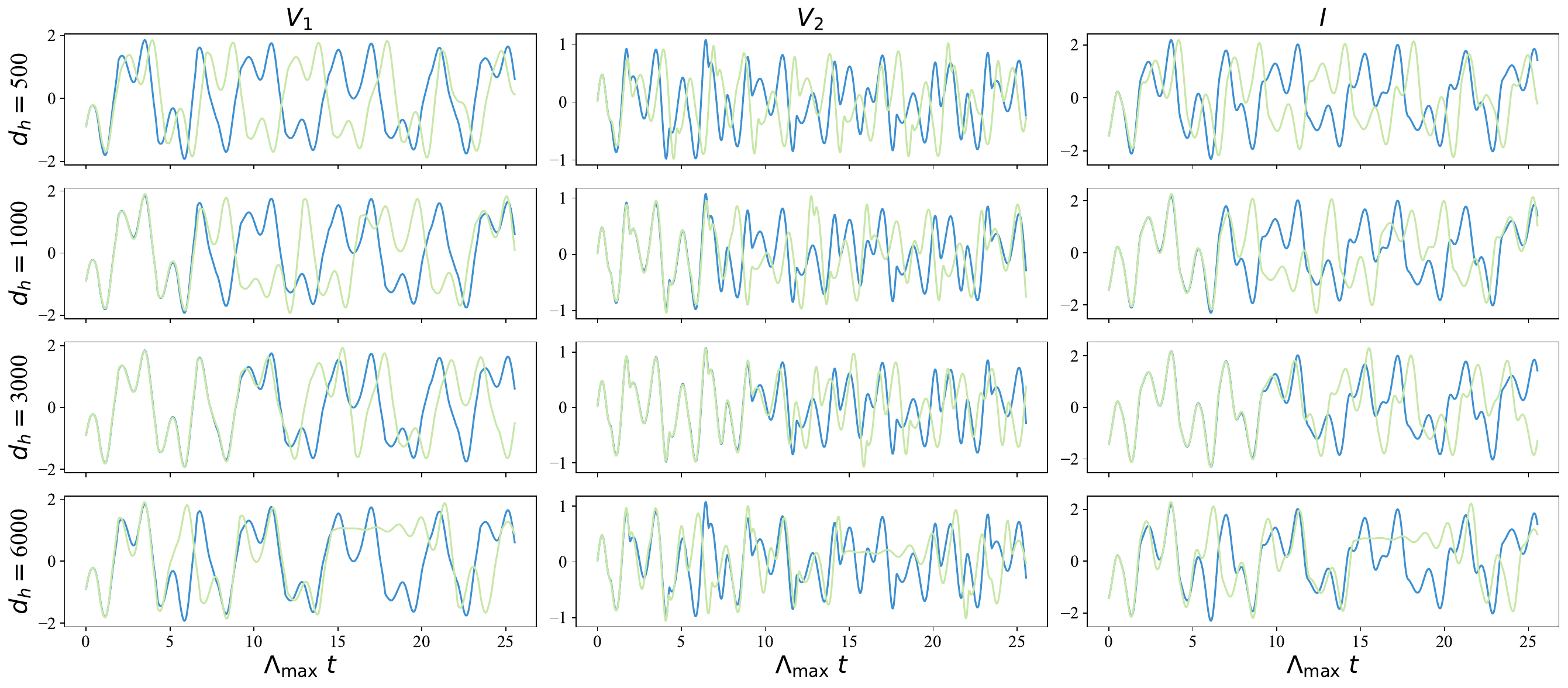}
    \caption{\textbf{Forecasting performance of RC with different reservoir dimensions on the double-scroll system.}
    Each row corresponds to a different reservoir dimension \(d_h\), and each column represents one state variable of the double-scroll system.
    The trajectory colors follow the same convention as in the main-text double-scroll experiment.}
    \label{fig:RC_DoubleScroll_different_sizes}
\end{figure}

\begin{figure}[!htbp]
    \centering 
    \fitfigure{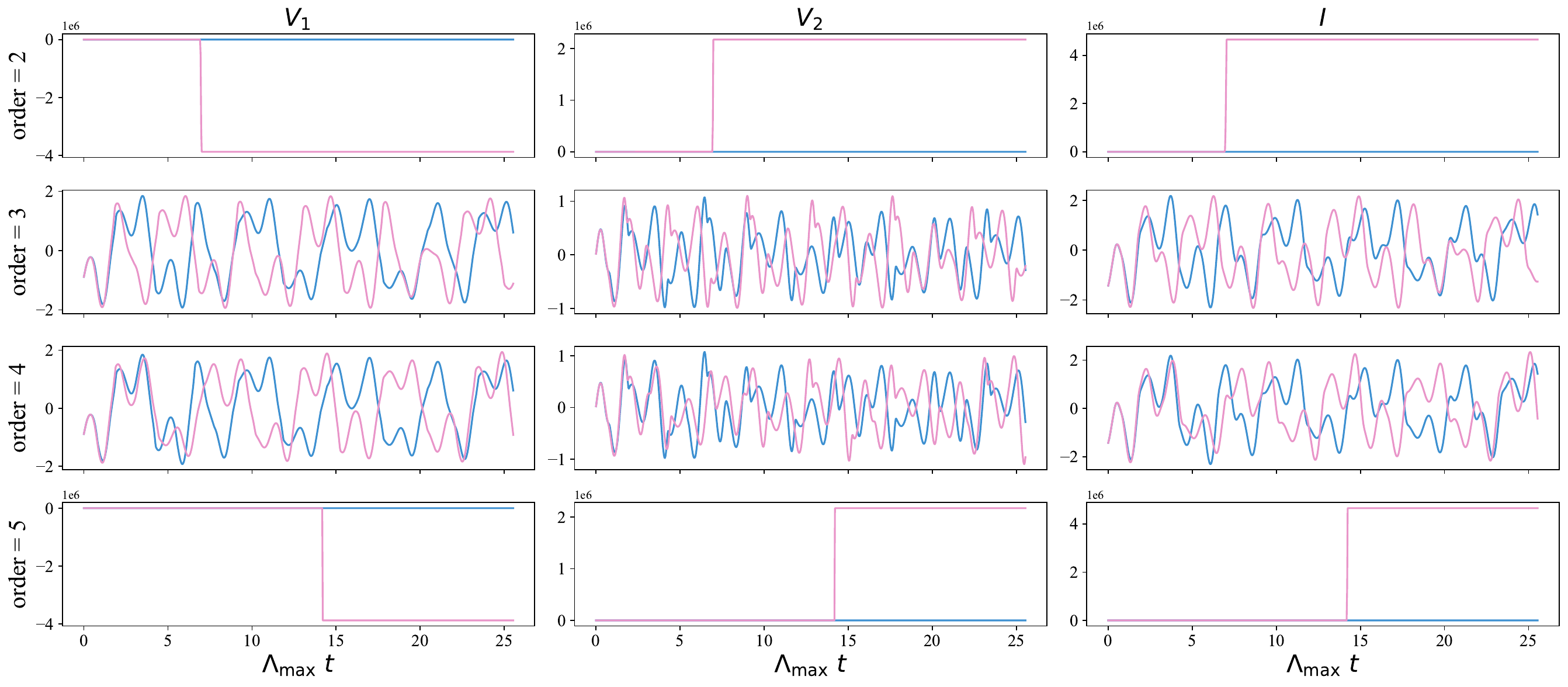}
    \caption{\textbf{Forecasting performance of NG-RC with different orders on the double-scroll system.}
    Rows correspond to different orders and columns represent the three state variables \(V_1\), \(V_2\) and \(I\).
    The trajectory colors follow the same convention as in the main-text double-scroll experiment.}
    \label{fig:NGRC_DoubleScroll_different_orders}
\end{figure}

\begin{figure}[!htbp]
    \centering
    \fitfigure{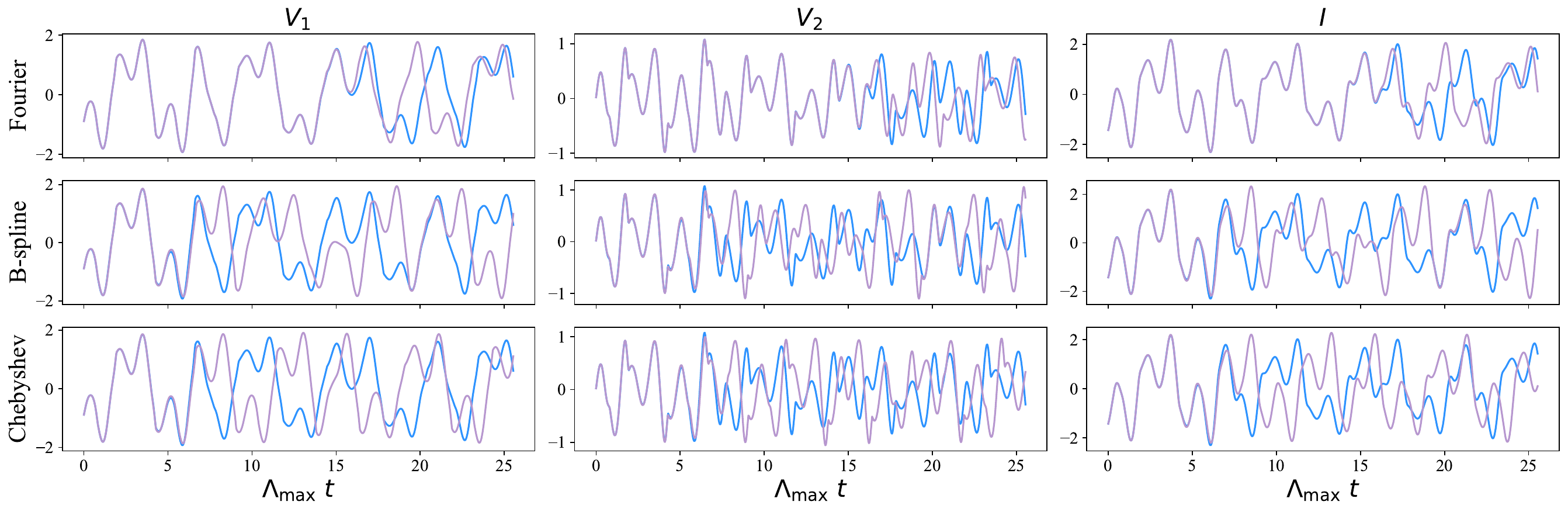}
    \caption{\textbf{Forecasting performance of KARC with different bases on the double-scroll system.}
    Rows correspond to different basis functions, including Fourier, B-spline and Chebyshev bases, and columns represent the three state variables \(V_1\), \(V_2\) and \(I\).
    The trajectory colors follow the same convention as in the main-text double-scroll experiment.}
    \label{fig:KARC_DoubleScroll}
\end{figure}

We first examine the sensitivity of RC to reservoir dimension on the double-scroll system. 
As shown in Figure~\ref{fig:RC_DoubleScroll_different_sizes}, enlarging the reservoir dimension need not yield a monotonic gain in prediction accuracy. While a larger reservoir offers a higher-dimensional representation space, the forecasts remain sensitive to the specific choice of \(d_h\); excessively large reservoirs often fail to ensure more accurate long-term rollouts. 
This observation aligns with the known hyperparameter sensitivity of RC, for which performance is critically dependent on the reservoir design and initialization.

Figure~\ref{fig:NGRC_DoubleScroll_different_orders} illustrates the sensitivity of NG-RC to the polynomial expansion order.
Our results reveal that NG-RC is highly susceptible to this choice.
Setting the order to two leads to severe numerical instability, causing the predicted trajectories to diverge rapidly from the ground truth.
A similar divergence is observed at order five, indicating that simply raising the expansion order need not improve forecasting and may instead compromise numerical stability. 
By contrast, third- and fourth-order models produce comparatively stable forecasts, with comparable performance over the tested horizon. 
Given that the fourth-order expansion entails a substantially larger feature dimension and higher computational cost without a clear accuracy benefit, we adopt the third-order configuration as the most efficient and stable choice for the main-text comparisons.

Figure~\ref{fig:KARC_DoubleScroll} compares the forecasting performance of KARC using different basis families. 
Among the tested options, the Fourier basis achieves the best performance, maintaining the closest agreement with the reference trajectories over the longest rollout. 
In contrast, while B-spline and Chebyshev bases still capture the overall oscillatory dynamics, their forecasts tend to deviate earlier.
Their performance is comparable to the best NG-RC results, though slightly weaker than the optimal RC configuration tested on this system.
Nevertheless, these results highlight that the choice of basis function is critical in KARC, with the Fourier basis being particularly well-suited to the oscillatory nature of the double-scroll dynamics.
\clearpage

\subsection{Kuramoto-Sivashinsky equation}
\begin{figure}[!htbp]
    \centering
    \fitfigure{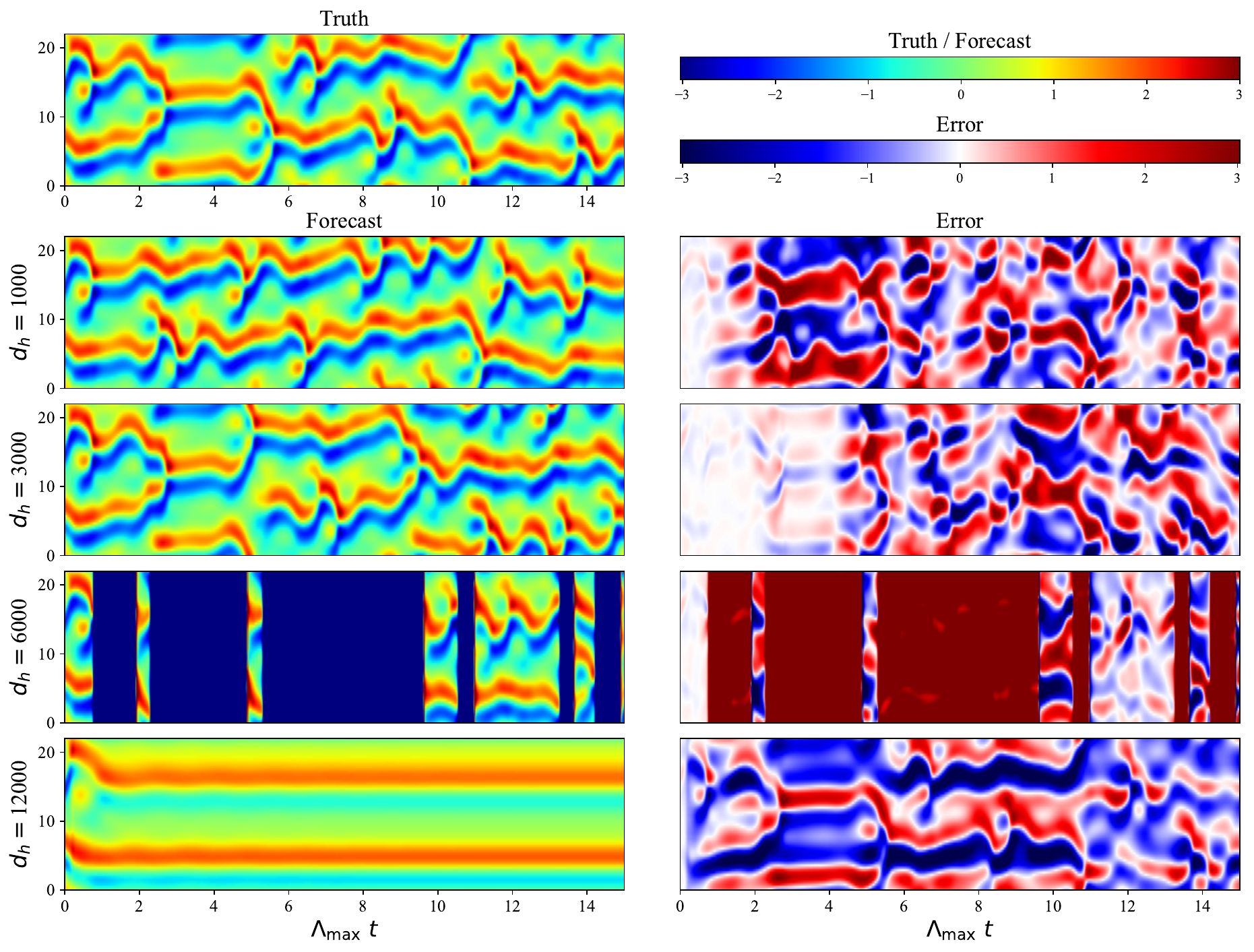}
    \caption{\textbf{Forecasting performance of RC with different reservoir dimensions on the Kuramoto-Sivashinsky equation.}
    The top-left panel shows the reference spatiotemporal field, and the subsequent rows show RC forecasts obtained with different reservoir dimensions \(d_h=1000,3000,8000\) and \(12000\). The right column reports the corresponding pointwise prediction errors. 
    }
    \label{fig:RC_KS_different_size}
\end{figure}

\begin{figure}[!htbp]
    \centering
    \fitfigure{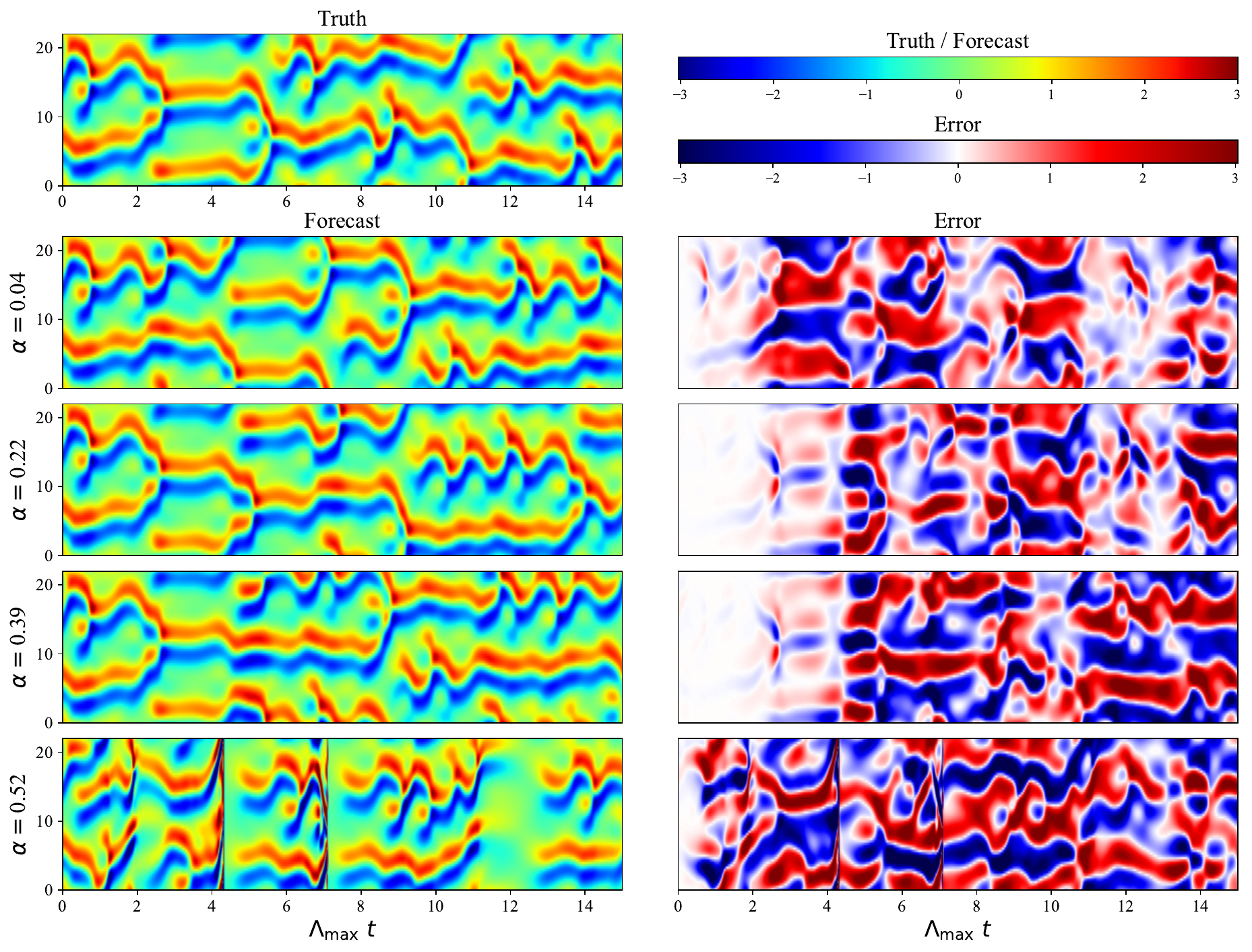}
    \caption{\textbf{Forecasting performance of VolterraRC with different kernel parameters \(\alpha\) on the Kuramoto-Sivashinsky equation.}
    The top row shows the reference spatiotemporal field, and the subsequent rows show the forecasts and corresponding pointwise errors for different \(\alpha\).
    }
    \label{fig:VolterraRC_KS_22}
\end{figure}

\begin{figure}[!htbp]
    \centering
    \fitfigure{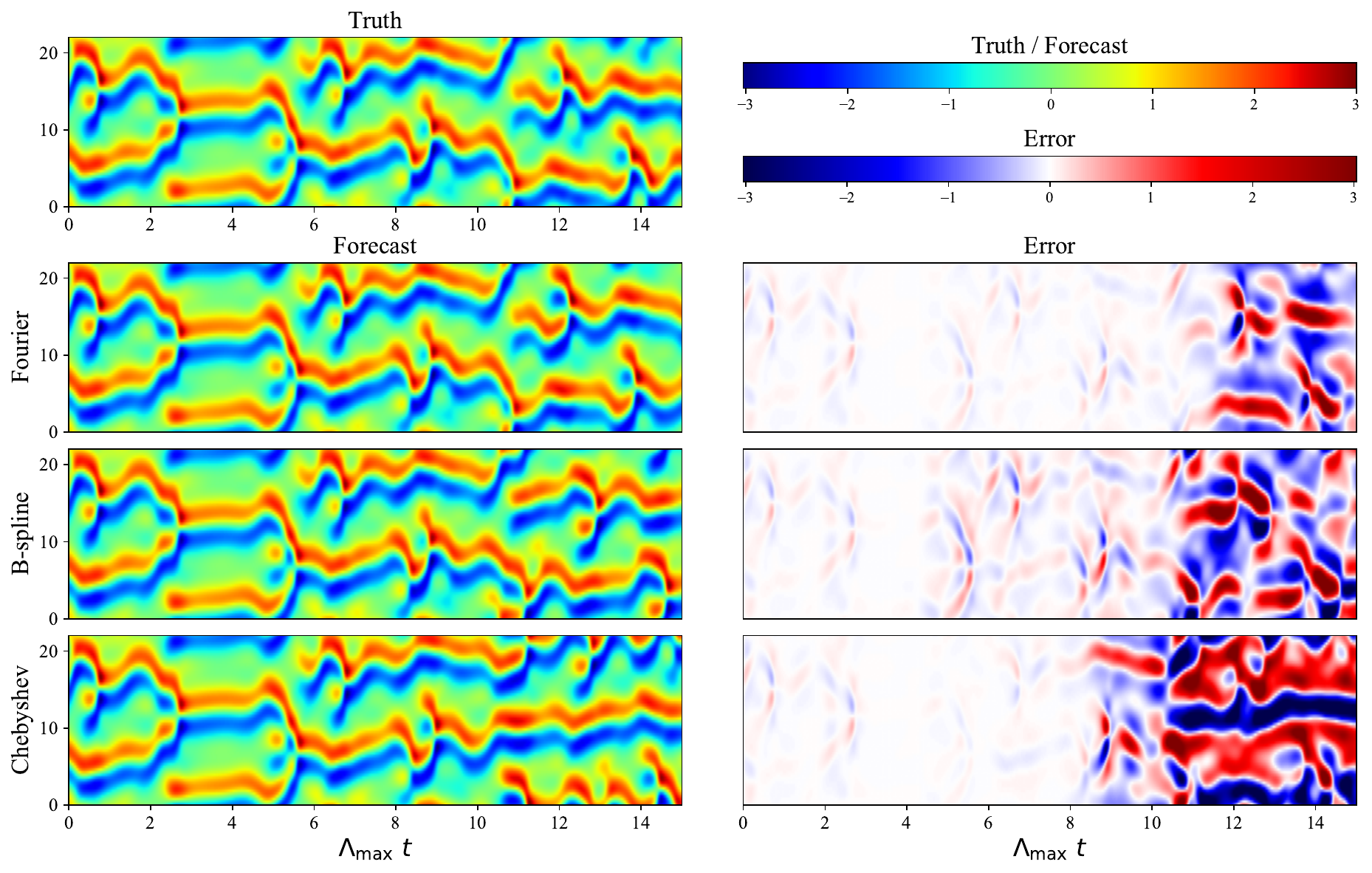}
    \caption{\textbf{Forecasting performance of KARC with different bases on the Kuramoto-Sivashinsky equation.}
    The top-left panel shows the reference spatiotemporal field, and the subsequent rows show KARC forecasts obtained using Fourier, B-spline and Chebyshev bases. The right column presents the corresponding pointwise prediction errors. The results compare how the choice of basis function affects the spatiotemporal forecasting accuracy of KARC.
    }
    \label{fig:KARC_KS_22}
\end{figure}

We next extend the sensitivity analysis to the Kuramoto-Sivashinsky equation. 
In Figure~\ref{fig:RC_KS_different_size}, the trend observed on the double-scroll system persists: increasing the reservoir dimension need not improve forecasting performance. 
Although a larger reservoir is conventionally expected to enhance expressivity, our results show no monotonic gain as \(d_h\) increases.
In particular, \(d_h=3000\) yields a comparatively stable forecast, whereas further enlargement degrades or destabilizes predictions. 
At \(d_h=12000\) , the predicted field becomes overly smooth, losing the characteristic chaotic structures of the reference solution. 
These findings confirm that RC performance is not governed by reservoir dimension alone, but remains sensitive to reservoir construction and hyperparameter choices.
 
Figure~\ref{fig:VolterraRC_KS_22} illustrates the sensitivity of VolterraRC to the kernel memory parameter \(\alpha\) on the Kuramoto-Sivashinsky equation. 
In the Volterra kernel, \(\alpha\) controls the contribution of previous temporal information: a larger \(\alpha\) corresponds to weaker decay of historical information and therefore retains longer memory from past states.
However, increasing \(\alpha\) does not yield monotonic improvement: while moderate values preserve the dominant spatiotemporal structures, pushing \(\alpha\) to \(0.52\) accelerates the loss of coherence and amplifies errors. 
This indicates that longer memory is not universally beneficial; the kernel parameter must instead be calibrated to balance historical information with long-term stability.

Finally, Figure~\ref{fig:KARC_KS_22} compares KARC with Fourier, B-spline, and Chebyshev bases on the Kuramoto-Sivashinsky equation. 
All three achieve notably long horizons, with Fourier reaching approximately 11 Lyapunov times, followed by B-spline and Chebyshev at about 10 and 9, respectively. 
These results show that KARC's effectiveness is not confined to a single basis family, though performance varies with the choice. 
Given its longest horizon and natural alignment with periodic boundary conditions, we adopt the Fourier basis as the default in the main-text experiments unless otherwise stated.
\clearpage

\subsection{Shallow Water equations}
\begin{figure}[!htbp]
    \centering
    \fitfigure{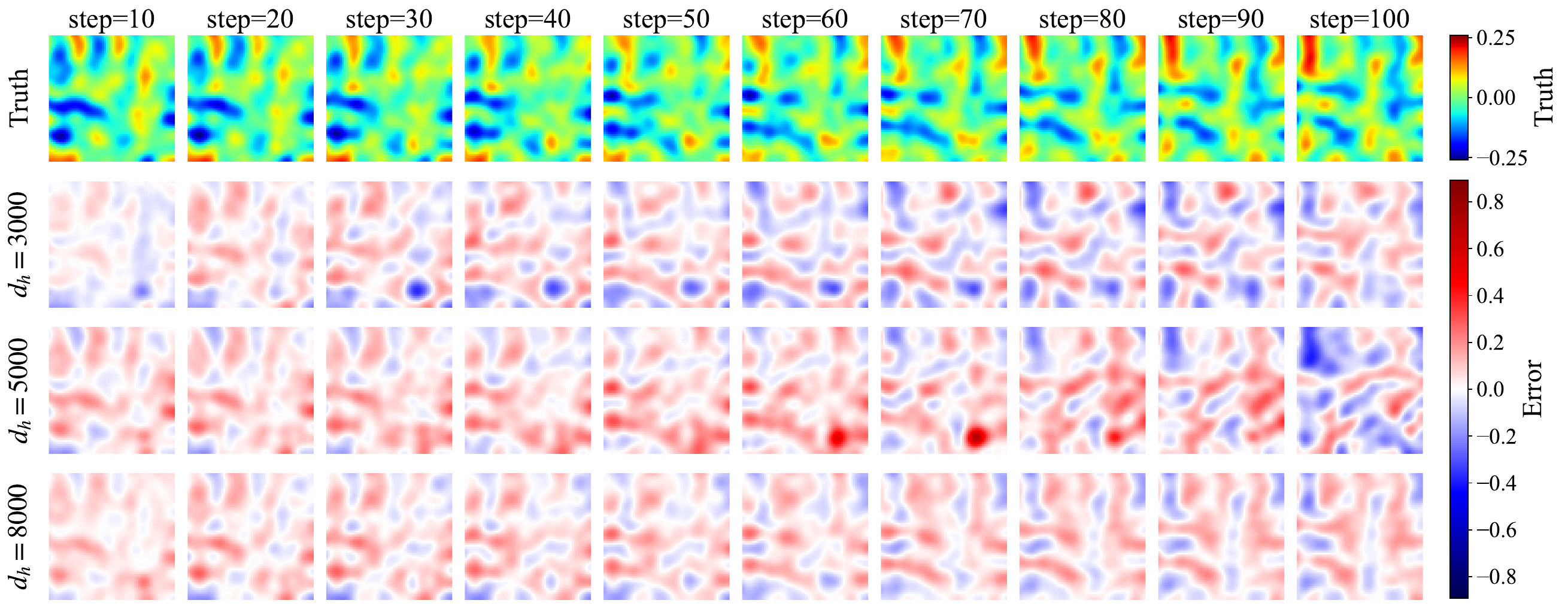}
    \caption{\textbf{Forecasting performance of RC with different reservoir dimensions on the shallow water equations.} 
    The first row shows the reference solution fields from step \(10\) to step \(100\), and the remaining rows show the corresponding prediction errors of RC with reservoir dimensions \(d_h=3000\), \(d_h=5000\), and \(d_h=8000\).}
    \label{fig:RC_SWE}
\end{figure}

\begin{figure}[!htbp]
    \centering
    \fitfigure{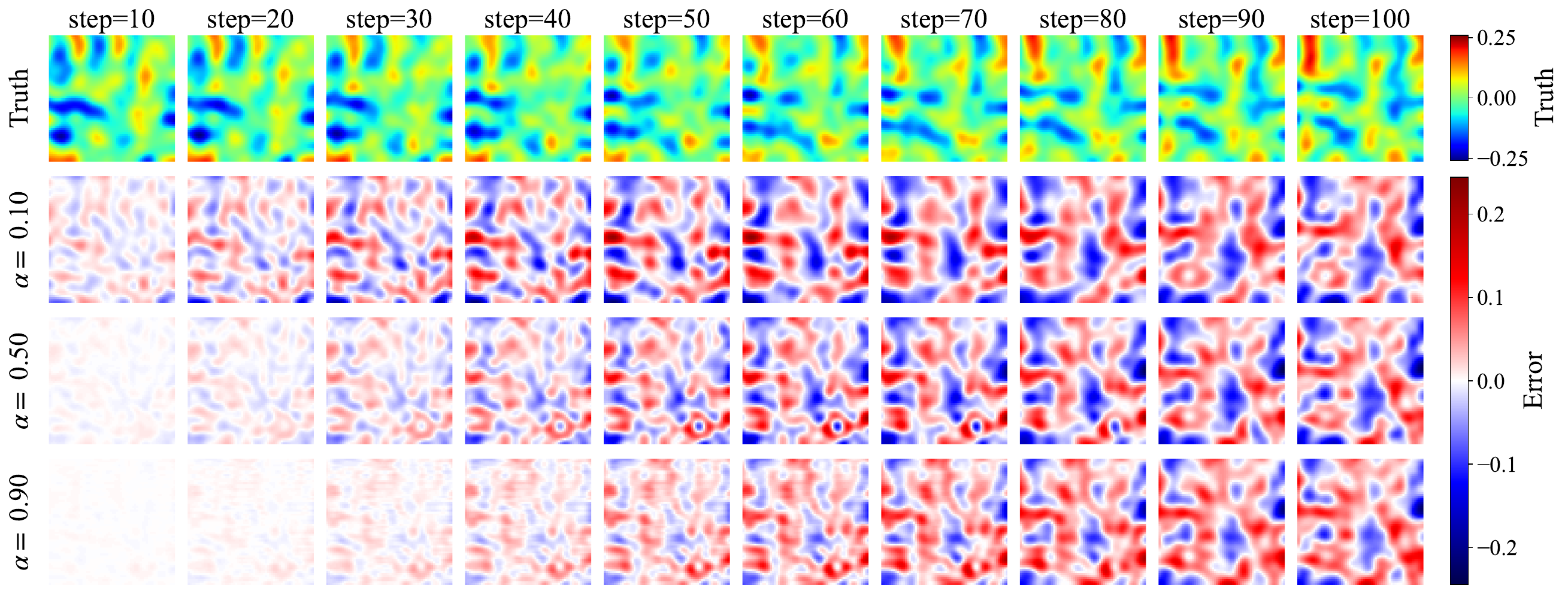}
    \caption{\textbf{Forecasting performance of VolterraRC with different kernel parameters \(\alpha\) on the shallow water equations.}
    The first row shows the reference solution fields from step \(10\) to step \(100\).
    The remaining rows show the corresponding pointwise prediction errors of VolterraRC with different values of the kernel memory parameter \(\alpha\).}
    \label{fig:VolterraRC_SWE}
\end{figure}

\begin{figure}[!htbp]
    \centering
    \fitfigure{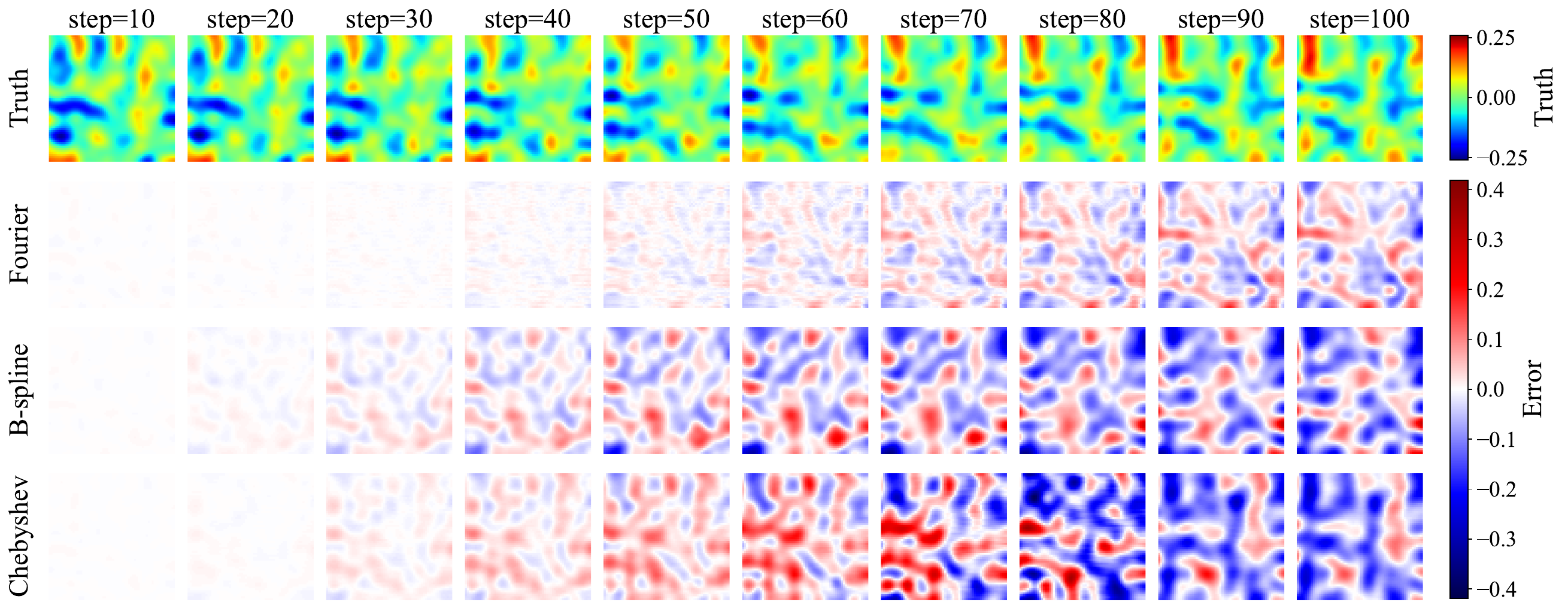}
    \caption{\textbf{Forecasting performance of KARC with different bases on the shallow water equation.} 
    The first row shows the reference solution fields from step \(10\) to step \(100\), and the remaining rows show the corresponding prediction errors of KARC using Fourier, B-spline, and Chebyshev basis functions.}
    \label{fig:KARC_SWE}
\end{figure}

We next turn to the shallow water equations, a benchmark that embodies additional physical constraints such as mass conservation. Figure~\ref{fig:RC_SWE} evaluates RC with varying reservoir dimensions.
Increasing \(d_h\) from 3000 to 8000 fails to visibly extend the prediction horizon; although larger reservoirs introduce more internal degrees of freedom, the error maps remain comparable across the tested dimensions, with substantial accumulation still observed during long-term rollout. 
These results confirm that RC performance on the shallow water equations is not dictated by reservoir dimension alone, but remains contingent on the overall construction and hyperparameter selection.

Figure~\ref{fig:VolterraRC_SWE} further evaluates the sensitivity of VolterraRC to the kernel memory parameter \(\alpha\).
Different from the results on the Kuramoto-Sivashinsky equation, increasing \(\alpha\) improves the forecasting performance on the shallow water equations. 
When \(\alpha=0.1\), the prediction error grows rapidly over the rollout horizon.
As \(\alpha\) increases to \(0.50\) and \(0.90\), the accumulated error is noticeably reduced, and the model better preserves the spatiotemporal evolution of the reference solution.
These results suggest that the shallow-water dynamics benefit from a longer effective memory in the Volterra kernel, while also highlighting that the optimal choice of \(\alpha\) is system dependent.

Figure~\ref{fig:KARC_SWE} compares KARC with different basis functions on the shallow water equations. 
Among the three, Fourier achieves the best long-term rollout, with substantially lower error accumulation than B-spline and Chebyshev, likely due to its ability to represent the dominant large-scale spatial modes in this periodic setting. 
Nevertheless, KARC with B-spline and Chebyshev still outperforms RC, confirming that the advantage of KARC is not attributable solely to the Fourier basis but rather to the broader structured basis-expansion framework.

\clearpage

\section{Additional text-to-image examples}

\begin{figure}[!htbp]
    \centering
    \fitfigure{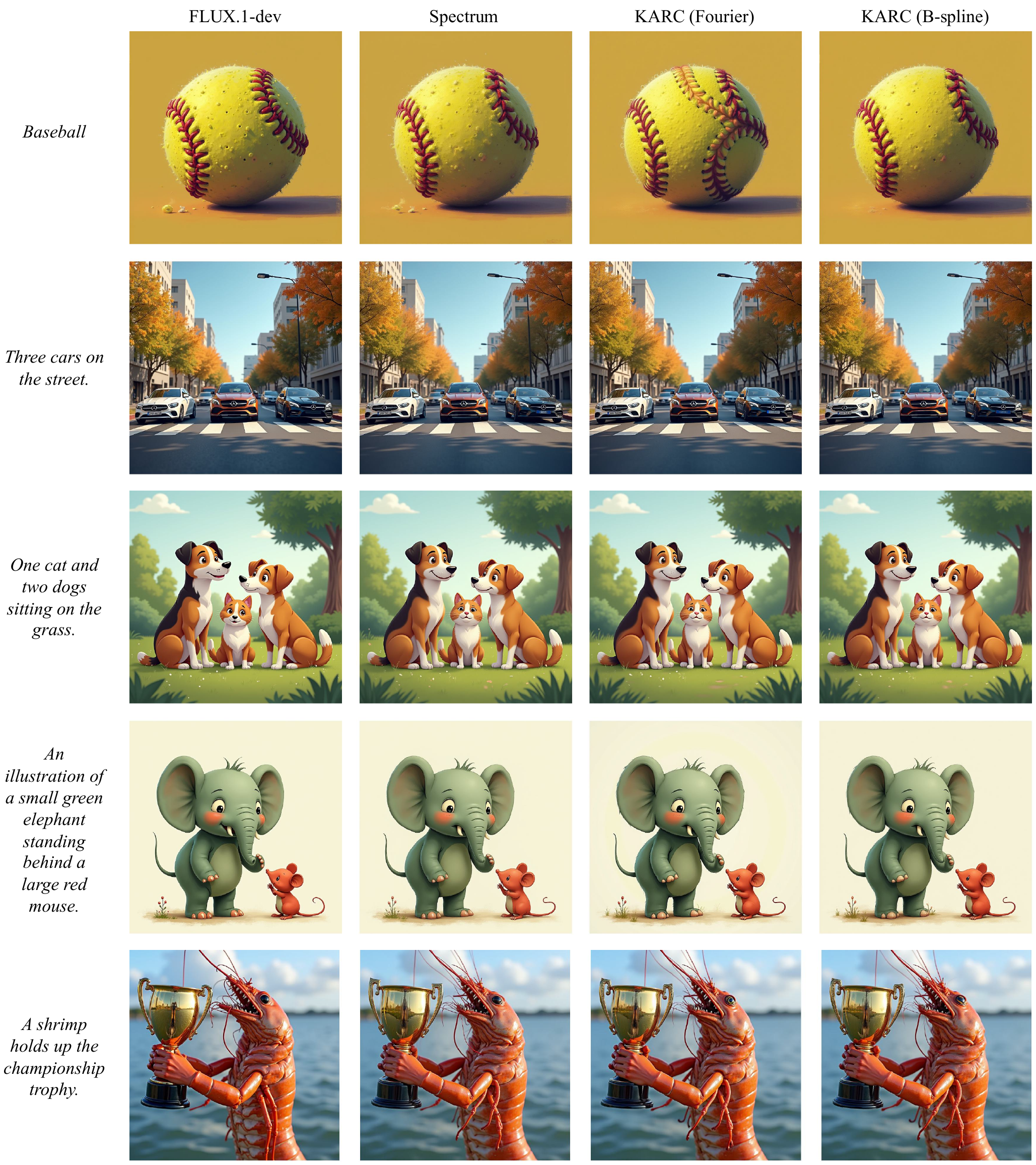}
    \caption{\textbf{Additional text-to-image generation examples.}  
    Each row corresponds to a text prompt. Each column shows the results produced by FLUX.1-dev (baseline), Spectrum, KARC with Fourier bases, and KARC with B-spline bases, respectively.}
    \label{fig:spectrum_fig}
\end{figure}

We further provide additional text-to-image generation examples using FLUX.1-dev~\cite{Flux} to evaluate the visual quality of KARC-based acceleration.
We compare the original Spectrum~\cite{Han2026Adaptive} acceleration method with KARC variants using Fourier and B-spline basis functions.
For each prompt, the original FLUX.1-dev output is used as the reference, and the accelerated results are compared in terms of semantic consistency and visual fidelity.
In Figure~\ref{fig:spectrum_fig}, both KARC-Fourier and KARC-B-spline preserve the main image content and overall composition of the reference generations.
Although minor differences appear in local textures and fine details, the generated images remain visually consistent with the prompts. 
These qualitative examples illustrate that KARC can be incorporated into the Spectrum acceleration framework without visibly disrupting the main semantic content or image composition.

\section{Applications to more examples}

\subsection{Lorenz63 system}

\begin{figure}[!htbp]
    \centering
    \fitfigure{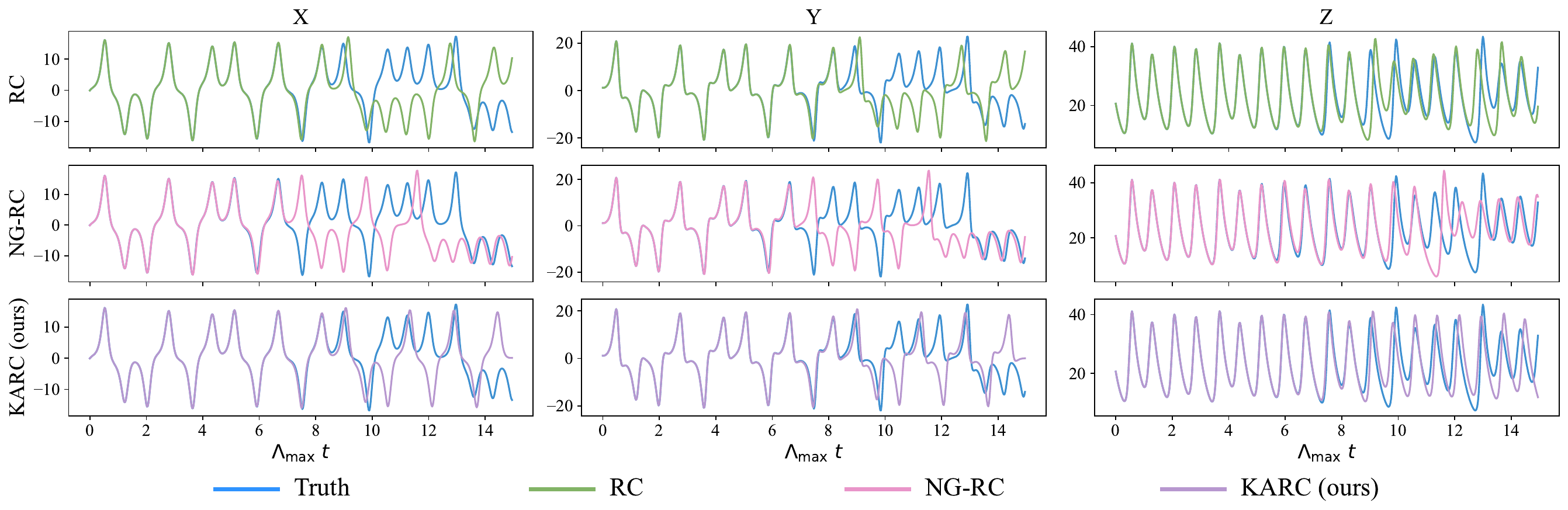}
    \caption{\textbf{Forecasting performance of RC, NG-RC, and KARC on the Lorenz63 system.} Rows correspond to different models, and columns correspond to the three state variables of the Lorenz63 system. $\Lambda_{\max}$ denotes the largest Lyapunov exponent, and one unit on the horizontal axis represents one Lyapunov time.}
    \label{fig:Lorenz_Comparison}
\end{figure}

In the main text, the double-scroll system was used as a representative low-dimensional chaotic ODE benchmark to assess the forecasting ability of KARC.
As an additional validation, we consider the Lorenz63 system, another classical low-dimensional chaotic system widely used in nonlinear dynamics and data-driven forecasting. 
The Lorenz63 system is governed by the following three-dimensional nonlinear ordinary differential equations:
\begin{align}
    \dot{x} &= \sigma (y-x), \\
    \dot{y} &= x(\rho-z)-y, \\
    \dot{z} &= xy-\beta z,
\end{align}
where \((x,y,z)\in\mathbb{R}^{3}\) denotes the system state and we use the standard chaotic parameter setting \( \sigma = 10, \rho = 28,\beta = \frac{8}{3}.\)
In this experiment, the training trajectory consists of 4000 data points, sampled at a rate of 40 points per unit time. The Lyapunov time of the system is estimated to be \(T_{\mathrm{LT}}=1.104\).

\begin{table}[!htbp]
    \centering
    \setlength{\tabcolsep}{10pt}
    \renewcommand{\arraystretch}{1.25}
    \begin{tabular}{lcccc}
        \toprule
        Model & Dimension & Total Time (s) \(\downarrow\) & NRMSE$@$1LT \(\downarrow\)& VPT ($\epsilon=0.1$) [LT] \(\uparrow\)\\
        \midrule
        RC & $500$ & $0.288$ & $1.529 \times 10^{-3}$ & $8.062$ \\
        NG-RC & $28$ & $0.151 $ & $1.061 \times 10^{-3}$ & $6.658$ \\ 
        KARC (ours) & $495$ & $0.188$ & $6.559 \times 10^{-4}$ & $8.152$ \\
        \bottomrule
    \end{tabular}
    \caption{\textbf{Quantitative comparison of forecasting performance on the Lorenz63 system.} The evaluation metrics are the same as those used in the previous experiments}
    \label{tab:Lorenz_table}
\end{table}

Following the same evaluation protocol as in the double-scroll experiment, we compare three models on the Lorenz63 system: RC, NG-RC, and KARC. 
Specifically, RC uses a reservoir dimension of 500, NG-RC adopts second-order polynomial features, and KARC uses a second-order feature construction with Chebyshev basis functions. 
In Figure~\ref{fig:Lorenz_Comparison}, all three models achieve comparable short-term forecasting performance on the Lorenz63 system. 
The NG-RC prediction starts to deviate from the reference trajectory at approximately the seventh Lyapunov time, whereas RC and KARC remain close to the true trajectory until around the eighth Lyapunov time. 
These results indicate that, on this low-dimensional chaotic ODE benchmark, KARC achieves forecasting accuracy comparable to RC and NG-RC, while retaining a deterministic feature construction and closed-form training.

Although the predicted trajectories of RC and KARC appear similar in Figure~\ref{fig:Lorenz_Comparison}, the quantitative results in Table~\ref{tab:Lorenz_table} reveal further differences. 
KARC achieves a shorter total time than RC, reducing the training cost from \(0.288\) s to \(0.188\) s. 
It also obtains a lower prediction error within the first Lyapunov time, with an NRMSE of \(6.559\times 10^{-4}\), compared with \(1.529\times 10^{-3}\) for RC. 
The table also further confirms that KARC outperforms NG-RC in terms of forecasting accuracy and prediction horizon. 
Although NG-RC has a smaller feature dimension and a shorter total time in this low-dimensional setting, its autonomous rollout deviates earlier from the reference trajectory.
These results show that KARC provides more accurate and more stable forecasting performance on the Lorenz63 system.

\subsection{Navier-Stokes equation}
\label{sec:KARC_FNO}

\begin{figure}[!htbp]
    \centering
    \fitfigure{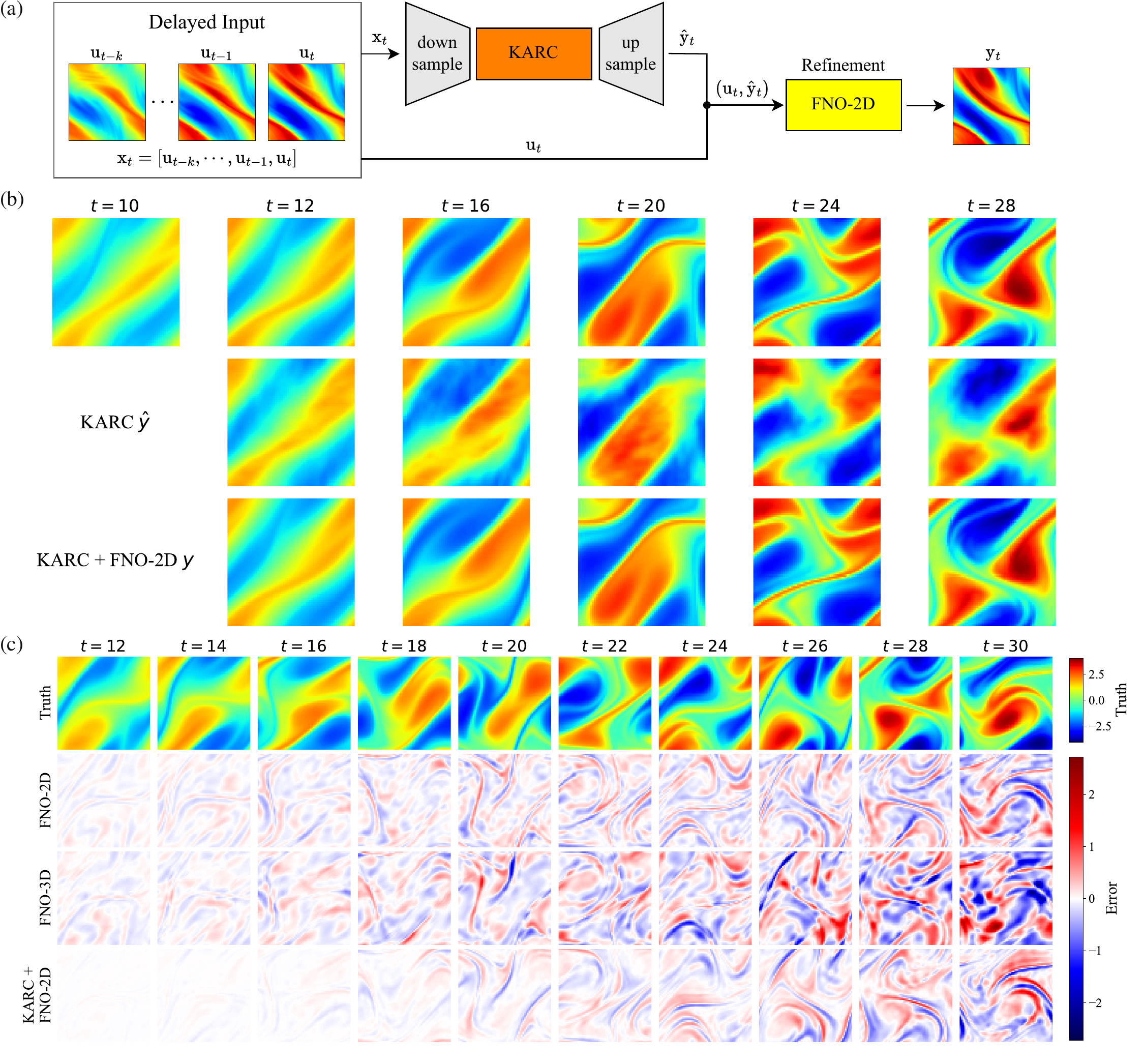}
    \caption{(a) Schematic illustration of the hybrid KARC+FNO-2D architecture.
    (b) Qualitative comparison between KARC-only and KARC+FNO-2D predictions at selected forecasting times. The rows show the $64\times64$ ground truth, the KARC-only prediction computed at $32\times32$ resolution and upsampled to $64\times64$, and the final $64\times64$ KARC+FNO-2D prediction, respectively.
    (c) Forecasting errors on the Navier-Stokes equation.The first row shows the reference vorticity fields at different prediction times, while the remaining rows show the corresponding prediction errors of FNO-2D, FNO-3D, and KARC+FNO-2D.
    }
    
    \label{fig:NS_comparison}
\end{figure}

To explore the potential of combining KARC with Fourier Neural Operator (FNO)~\cite{FNO}, we further conduct experiments on the Navier-Stokes equation. 
Unlike the low-dimensional ODE systems considered above, the Navier-Stokes equation describes high-dimensional spatiotemporal fluid dynamics and provides a challenging benchmark for multiscale turbulent dynamics.
We consider the incompressible vorticity formulation:
\begin{align}
    \partial_t w(x,t) + u(x,t)\cdot \nabla w(x,t) &= \nu \Delta w(x,t) + f(x),
    && x\in(0,1)^2,\; t\in(0,T], \\
    \nabla \cdot u(x,t) &= 0,
    && x\in(0,1)^2,\; t\in[0,T], \\
    w(x,0) &= w_0(x),
    && x\in(0,1)^2 .
\end{align}
Here, $w(x,t)$ denotes the vorticity field, $u(x,t)$ is the incompressible velocity field, $\nu$ is the kinematic viscosity, $f(x)$ is the external forcing term, and $w_0(x)$ is the initial vorticity field.
In this experiment, we consider a viscosity coefficient of \(\nu = 1\times 10^{-4}\).
The training set contains 1000 solution samples. For each sample, the model observes the vorticity field over the time interval \((0,10]\) and is trained to predict its future evolution over the interval \((10,30]\).

We propose a hybrid KARC+FNO-2D architecture, in which KARC serves as a coarse predictor for capturing the large-scale flow structures, while FNO-2D acts as a refinement module to recover small-scale filamentary vortices. 
The overall architecture is illustrated in Figure~\ref{fig:NS_comparison}(a).
A delayed input sequence is first processed by KARC on a downsampled grid, producing a coarse forecast that captures the dominant large-scale evolution of the vorticity field. 
This coarse prediction is then upsampled and concatenated with the current state as the input to FNO-2D, which refines the spatial details and generates the final high-resolution forecast.

Figure~\ref{fig:NS_comparison}(b) presents a qualitative comparison of the forecasting results. 
KARC alone captures the main large-scale flow patterns, but tends to miss fine-scale filamentary structures.
By contrast, the FNO-2D refinement stage improves the reconstruction of these small-scale vortices and produces a more detailed prediction. 
Importantly, since KARC operates on a downsampled grid, it does not require high-resolution inputs for feature construction.
This reduces the computational cost of the KARC stage, while the subsequent FNO-2D refinement compensates for the loss of small-scale information introduced by downsampling.

Figure~\ref{fig:NS_comparison}(c) shows the forecasting results of FNO-2D, FNO-3D, and the proposed KARC+FNO-2D hybrid architecture for the Navier-Stokes equation with viscosity \(\nu=1\times10^{-4}\).
As the forecast horizon increases, both FNO-2D and FNO-3D gradually accumulate visible errors, especially in regions with complex vortical and filamentary structures.
In contrast, KARC+FNO-2D produces weaker error patterns over most time steps, indicating that the coarse prediction provided by KARC helps preserve the large-scale flow evolution, while the FNO-2D refinement module further improves the reconstruction of fine-scale structures.
These results suggest that the hybrid architecture provides a promising way to combine the efficient temporal feature construction of KARC with the spatial refinement capability of FNO-2D for high-dimensional fluid forecasting.

\clearpage
\bibliography{bib}